\documentclass[twocolumn,journal]{IEEEtran}
\usepackage[square, comma, sort&compress, numbers]{natbib}
\usepackage{epsfig}
\usepackage{epstopdf}
\usepackage{graphicx}
\usepackage{citesort}
\usepackage{amssymb}
\usepackage{amsmath}
\usepackage{color}
\usepackage{url}
\usepackage{array}
\usepackage{multirow}
\usepackage{algorithm}
\usepackage{algorithmic}
\usepackage[table]{xcolor}
\usepackage{bm}
\usepackage{tablefootnote}
\usepackage{threeparttable}
\usepackage{color}
\usepackage{colortbl}
\usepackage{mdwmath,mdwtab,amsmath}
\usepackage{autobreak}
\usepackage{amsmath,amssymb}
\usepackage{array}
\usepackage{multirow}
\usepackage{tabularx}
\usepackage{makecell}
\usepackage{graphicx}
\usepackage{threeparttable}
\usepackage{tablefootnote}
\newcolumntype{Y}{>{\centering\arraybackslash}X}

\newlength{\figurewidth}
\newlength{\smallfigurewidth}

\usepackage{tikz,xcolor,hyperref}
\definecolor{lime}{HTML}{A6CE39}
\DeclareRobustCommand{\orcidicon}{
	\begin{tikzpicture}
		\draw[lime, fill=lime] (0,0)
		circle[radius=0.16]
		node[white]{{\fontfamily{qag}\selectfont \tiny \.{I}D}}; 
	\end{tikzpicture}
	\hspace{-2mm}
}
\foreach \x in {A, ..., Z}{%
	\expandafter\xdef\csname orcid\x\endcsname{\noexpand\href{https://orcid.org/\csname orcidauthor\x\endcsname}{\noexpand\orcidicon}}
}

\hypersetup{hidelinks}

\begin{document}
	\title
	{LO-Det: Lightweight Oriented Object Detection in Remote Sensing Images
	}
	\author{%
		Zhanchao Huang\hspace{-1.5mm}\orcidA{},
		Wei Li,~\IEEEmembership{Senior Member,~IEEE},
		Xiang-Gen Xia,~\IEEEmembership{Fellow,~IEEE},  
		\\
		Hao Wang,
		Feiran Jie,
		and
		Ran Tao,~\IEEEmembership{Senior Member,~IEEE}
		\thanks{%
			This work was supported by the National Natural Science Foundation of China under Grant 61922013 and U1833203, and by the Beijing Natural Science Foundation under Grant L191004 and JQ20021. (Corresponding Author: Wei Li; e-mail: liwei089@ieee.org)}
		\thanks{%
			Zhanchao Huang, Wei Li, Hao Wang, Ran Tao are with the School of Information and Electronics, Beijing Institute of Technology, and Beijing Key Lab of Fractional Signals and Systems, 100081 Beijing, China. (e-mail: zhanchao.h@outlook.com; liwei089@ieee.org; haohaolalahao@icloud.com; rantao@bit.edu.cn).}
		\thanks{%
			Xiang-Gen Xia is with the Department of Electrical and Computer Engineering, University of Delaware, Newark, DE 19716, USA (e-mail: xxia@ee.udel.edu).}
		\thanks{%
			Feiran Jie is with the Luoyang Institute of Electro-Optical Equipment, The Aviation Industry Corporation of China, Ltd., 471000, Luoyang, Henan, China, (e-mail: bill.jfr@163.com).}
	}
\maketitle
\thispagestyle{empty}
\pagestyle{empty}

\begin{abstract}
A few lightweight convolutional neural network (CNN) models have been recently designed for remote sensing object detection (RSOD). However, most of them simply replace vanilla convolutions with stacked separable convolutions, which may not be efficient due to a lot of precision losses and may not be able to detect oriented bounding boxes (OBB). Also, the existing OBB detection methods are difficult to constrain the shape of objects predicted by CNNs accurately. In this paper, we propose an effective lightweight oriented object detector (LO-Det). Specifically, a channel separation-aggregation (CSA) structure is designed to simplify the complexity of stacked separable convolutions, and a dynamic receptive field (DRF) mechanism is developed to maintain high accuracy by customizing the convolution kernel and its perception range dynamically when reducing the network complexity. The CSA-DRF component optimizes efficiency while maintaining high accuracy. Then, a diagonal support constraint head (DSC-Head) component is designed to detect OBBs and constrain their shapes more accurately and stably. Extensive experiments on public datasets demonstrate that the proposed LO-Det can run very fast even on embedded devices with the competitive accuracy of detecting oriented objects.
\end{abstract}


\section{Introduction}
Benefitted from the rapid improvement of graphics processing performance and the easier availability of high-resolution remote sensing (RS) images, RSOD methods based on convolutional neural networks (CNNs) have attracted more attentions recently.

From universal detectors, such as Faster R-CNN \cite{renFasterRCNNRealTime2017a}, SSD \cite{liuSSDSingleShot2016a}, and YOLO \cite{redmonYOLOv3IncrementalImprovement2018}, etc., to dedicated detectors developed for RS scenes, the design of CNN-based detectors is more and more in line with the requirements of RS tasks and the detection accuracy continues to increase. For example, Cheng et al. \cite{chengLearningRotationInvariantConvolutional2016} proposed a rotation-invariant CNN (RICNN) model to deal with the problem of object rotation variations. Yang et al. \cite{yangSCRDetMoreRobust2019} designed a rotation detector for small, cluttered and oriented objects which fused multi-layer feature and introduced attention mechanism for higher detection accuracy. Fu et al. \cite{fu2020point} developed a point-based estimator embedded in the region-based detector to improve the performance of detecting oriented objects in remote sensing images.

The existing improvements on CNNs for RSOD tasks can be summarized into three categories: the multi-scale feature enhancement, the visual attention mechanism, and the application of different pipelines. In terms of multi-scale feature enhancement \cite{huang2020dc}, the feature pyramid \cite{linFeaturePyramidNetworks2017a} and the inception structure \cite{szegedy2016rethinking} have always been research hotspots \cite{wangFMSSDFeatureMergedSingleShot2020,fuRotationawareMultiscaleConvolutional2020}. In terms of visual attention mechanism, various improvements based on SENet \cite{huSqueezeandExcitationNetworks2018} and CBAM \cite{wooCBAMConvolutionalBlock2018} to develop channel and spatial attentions have been continuing \cite{yangSCRDetMoreRobust2019,zhangCADNetContextAwareDetection2019,9364888}. In terms of detection pipelines, discussions from two-stage to one-stage \cite{yangSCRDetDetectingSmall2020b} and from anchor-based to anchor-free \cite{lin2019ienet} have never been stopped. These studies have promoted the improvement of detection accuracy in the initial rising stage when CNN was introduced into high-resolution RSOD. However, at present, researches in this field have changed from incremental development to stock improvement. The CNN models are becoming larger and more complex, and the complexity of them is becoming higher. But most of the existing researches in RSOD tasks may ignore the speed loss caused by using extremely deep backbone networks and the addition of various feature enhancements and visual attention mechanisms. Regrettably, the current development of graphics processing units (GPUs) has not yet reached the stage where computing power can be piled up regardless of cost, and not everyone can afford expensive GPUs, such as RTX 3090 or even Tesla V100. To deal with this problem, lightweight model design has been recently considered in the literature.

For lightweight CNN design, MobileNet \cite{howard2017mobilenets,sandler2018mobilenetv2,howard2019searching}, ShuffleNet \cite{zhang2018shufflenet,ma2018shufflenet}, GhostNet \cite{han2020ghostnet}, etc., have provided many good ideas although they are designed for image classification tasks. These networks are also applied to replace the backbone for object detection. But in addition to the backbone, a modern CNN-based detector usually includes two additional components, feature fusion module (neck) and prediction module (head). The number of convolutional layers of these components can even be comparable to that of the backbone, which has a great impact on the performance. However, the structure of stacked separable convolutions in these components used by most lightweight detectors is not efficient. Besides, RSOD tasks often have the demand to detect oriented bounding boxes (OBBs) in addition to horizontal bounding boxes (HBBs). These all need to be considered when designing a detector for RSOD tasks.

In summary, the lack of lightweight detectors developed for RSOD tasks, the inefficient design of stacked separable convolutions in detection neck component, and the inability of ordinary detection head component for OBB detection are all challenges in the current research. In this regard, this work provides a practical method of lightweight CNN design for RSOD tasks, which is rarely studied at present. The contributions of this work are summarized as follows:

1) A lightweight detector LO-Det for RSOD tasks is proposed for edge computing on embedded devices. This detector has comparable detection accuracy with the main existing detectors at a much faster inference speed and using much fewer computing resources.

2) A CSA structure is designed to simplify the complexity of stacked separable convolutions, and a DRF mechanism based on CSA is developed to customize the convolution kernel and its perception range dynamically for higher quality feature extraction and fusion. The proposed CSA-DRF component optimizes efficiency of the detector while maintaining high detection accuracy.

3) A DSC-Head is proposed to enable LO-Det for detecting OBB annotated objects. Furthermore, it makes the shape regression of OBBs more accurately by diagonal support constraint, and alleviates the boundary value problem of the existing OBB detection methods by the designed M-Sigmoid function.

The remainder of this paper is organized as follows. Section II reviews the related works and analyzes the existing problems. In Section III, a detailed description of the proposed LO-Det is presented. In Section IV, extensive experiments are conducted and the results are discussed. Conclusions are summarized in Section V.

\section{Related Works and Analysis}
\subsection{Lightweight CNNs}

With the development of CNNs, research on building small and efficient network models for embedded devices has become one of the important considerations in recent years\cite{wang2018pelee,yang2019legonet}. Xception \cite{chollet2017xception} uses the depth-wise convolution operation to improve Inceptionv3, and achieves better accuracy and efficiency under the condition of equivalent model parameters. SqueezeNet \cite{iandola2016squeezenet} compresses the number of model parameters by designing the bottleneck module, but the inference time is longer at the expense of network parallelism. In addition to improving VGGNet with depthwise convolution, MobileNet \cite{howard2017mobilenets} also designes a new activation function ReLU6 for lightweight networks, which can be more robust under low-precision calculations. MobileNetv2 \cite{sandler2018mobilenetv2} designs the inverted residual module with linear bottleneck based on MobileNet \cite{howard2017mobilenets}, which significantly improves the accuracy and speed of image classification. MobileNetv3 \cite{howard2019searching} optimizes the network structure parameters of MobileNet by neural architecture search (NAS) technology. ShuffleNet \cite{zhang2018shufflenet} designs the pointwise group convolution and channel shuffle modules for the computationally intensive problem of dense pointwise convolution, which makes its speed faster than MobileNet. ShuffleNetv2 \cite{ma2018shufflenet} analyzes the impact of memory access cost (MAC) on the inference speed of CNN and improves the performance of ShuffleNet. GhostNet \cite{han2020ghostnet} is a model with lower theoretical calculation complexity by applying a series of linear operations with low cost to extend feature maps.

In terms of object detection, the authors \cite{howard2017mobilenets} design an SSD-Lite detector by simply replacing the backbone of the SSD detector with MobileNet. Light-Head R-CNN \cite{liLightHeadRCNNDefense2017a} lights the head of the detector by using a sparse feature map before RoI Pooling. ThunderNet \cite{qin2019thundernet} is a compressed RPN sub-network for generating region proposals, and integrates local and global features through a context enhancement module to enhance feature expression. Lightdet \cite{tang2020lightdet} suggests a change in backbone structure as well as neck component to preserve more feature maps from shallow levels of the backbone.

In the field of RSOD, Ding et al. \cite{ding2018light} have made preliminary explorations and proposed a light and faster regional convolutional neural network. Although its speed is still slower than SSD \cite{liuSSDSingleShot2016a} and YOLO \cite{redmonYouOnlyLook2016a} and only evaluated the performance of detecting planes and cars, its detection accuracy is higher. Wang et al. \cite{wang2020ship} have developed a Lightweight CNN for ship detection in infrared images. The CNN model of this work uses a simple 4-layer convolution + pooling structure to makes the network lightweight, which is more suitable for the single-category object detection tasks, such as ship detection. As far as we know, most of the existing applications just simply replace the backbone of SDD \cite{liuSSDSingleShot2016a} or YOLO \cite{redmonYouOnlyLook2016a} with lightweight networks, such as MobileNet \cite{howard2017mobilenets}.

\subsection{Oriented Detection Head}

\begin{figure}[htbp]
	\centering
	\epsfig{width=0.45\textwidth,file=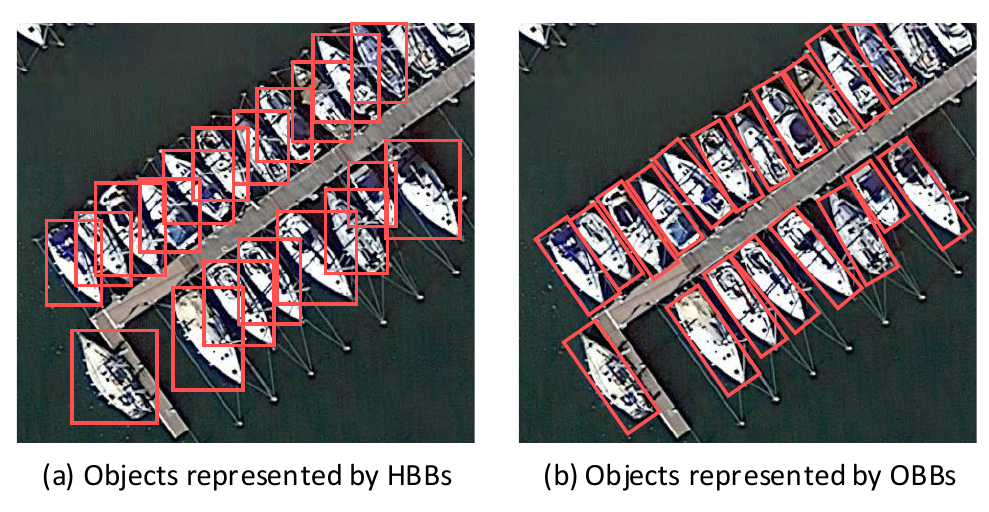}
	\caption{Objects represented by HBBs and OBBs.}\label{fig:1}
\end{figure}

Because objects in RS images usually have arbitrary-orientation characteristics, and some datasets, e.g., DOTA \cite{xiaDOTALargeScaleDataset2018} as shown in Fig.~\ref{fig:1}, also provide labels of oriented bounding box (OBB). The detection heads of many detectors have been improved to match this task. The methods with improvements of these detection head modules can be divided into two categories, angle-based methods and vertex-based methods. 

R$^{2}$CNN \cite{jiang2017r2cnn}, RRPN \cite{nabati2019rrpn}, SCRDet \cite{yangSCRDetMoreRobust2019}, FFA \cite{fuRotationawareMultiscaleConvolutional2020} and other detectors use angle-based OBB detection heads, which bring an additional angle parameter. However, the periodicity problem of angle regression and the inability to predict arbitrary-shaped quadrilaterals are the limitations of these methods.

Gliding Vertex \cite{xu2020gliding}, RSDet \cite{qian2019learning}, etc., use vertex-based OBB detection heads that can represent any quadrilaterals. These methods usually have higher detection accuracy compared with the angle-based method that can only represent the rotated rectangular bounding boxes. Among these methods, although Gliding Vertex \cite{xu2020gliding} avoids the vertex sorting and achieves the current best accuracy, too much freedom of vertex regression and boundary value problem may also lead to unstable results.

\subsection{Baseline, Problems, and Analysis}

As far as we know, there are very few researches on lightweight detectors specifically for RSOD tasks. Furthermore, most of these works are to directly replace the backbone of the universal detectors with lightweight backbone like MobileNet \cite{howard2017mobilenets} as a new lightweight detector. In order to analyze the problems of such a detector more comprehensively, and derive the motivation of the proposed solution accordingly, YOLOv3 \cite{redmonYOLOv3IncrementalImprovement2018} model, which has a very competitive advantage in detection speed and considerable accuracy, is improved by the simple replacement scheme mentioned above as the baseline for this work.

\begin{figure}[htbp]
	\centering
	\epsfig{width=0.47\textwidth,file=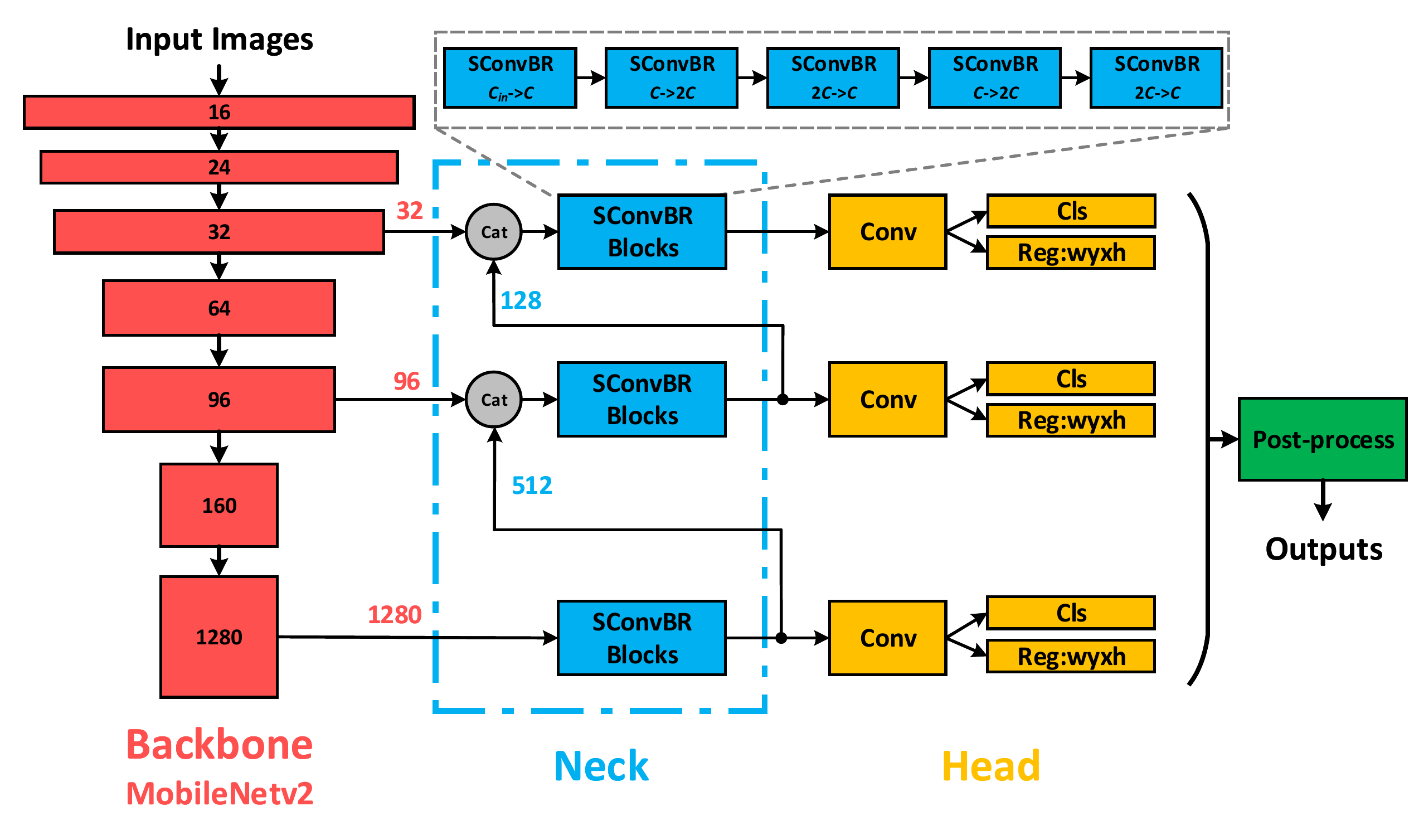}
	\caption{Structure of the baseline model (YOLOv3+MobileNetv2).}\label{fig:2}
\end{figure}

Specifically, MobileNetv2 \cite{sandler2018mobilenetv2} is adopted to replace the backbone of YOLOv3. Also, separable convolutions are employed instead of the vanilla convolutions in the neck component of YOLOv3. The structure of the baseline model is shown in Fig.~\ref{fig:2}. Among them, the red part is MobileNetv2, and the numbers in red blocks denote the number of feature maps output by each module of MobileNetv2. The blue part is the neck component composed of stacked separable convolutions. The yellow part is the head component for detecting HBBs. The green part indicates the post-process operation which refers to the non-maximum suppression (NMS) algorithm for filtering redundant bounding boxes, the conversion of the predicted coordinates into the coordinates on the output images, and the visual display.

This is the so-called lightweight improvement scheme of the most existing lightweight detectors, that is, choosing a general detector and replacing the original backbone with a lightweight backbone. But is it really efficient?

1) The inefficiency of stacked separable convolutions

\begin{figure}[htbp]
	\centering
	\epsfig{width=0.49\textwidth,file=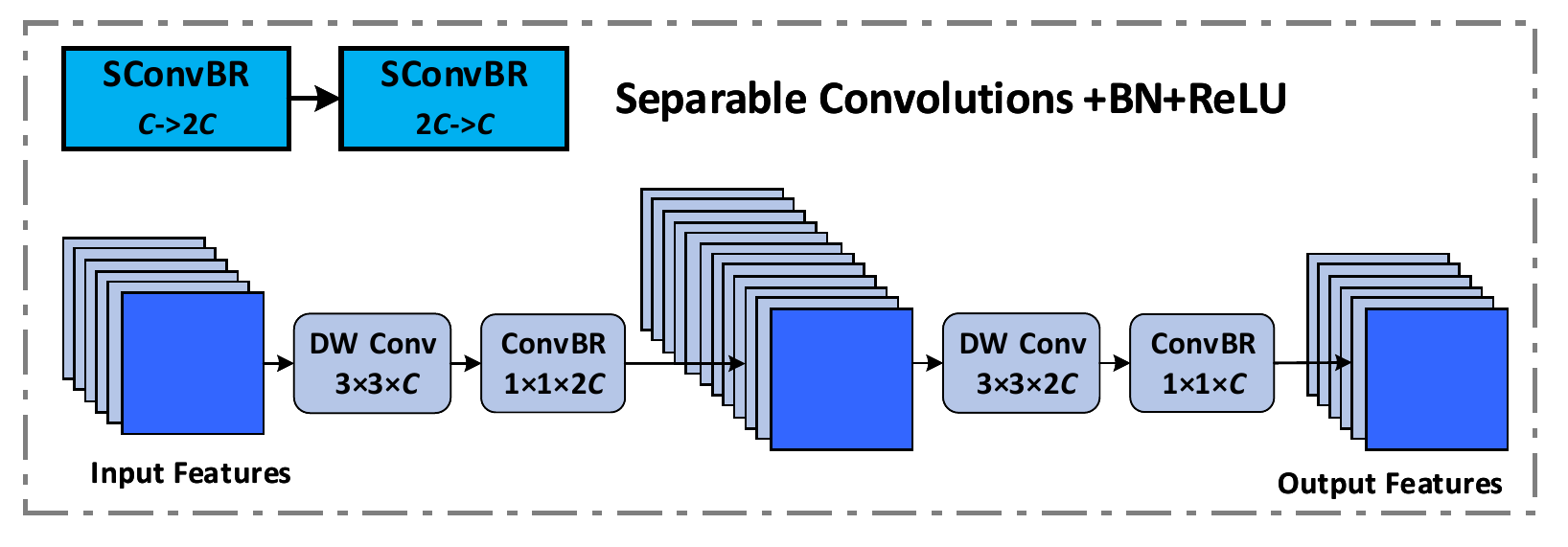}
	\caption{Structure of the SConv block.}\label{fig:3}
\end{figure}

It can be observed from Fig.~\ref{fig:2} that there are series of stacked separable convolutions (SConvs) used to fuse contextual features in the neck component. The detailed structure of these operations is shown in Fig.~\ref{fig:3}. Among them, each SConv consists of a 3×3 depth-wise convolution (DWConv) and a 1×1 vanilla convolution with nonlinear activation function used to adjust the number of feature maps (channels). Through these stacked SConv, the number of feature maps is frequently changing by a factor of 2 or 1/2. 

Take two SConvs as a group (SConv block) and calculate its floating-point operations (FLOPs) to represent the computational complexity. The dimensions of the input and output feature maps are defined as $W \times H \times C$, where $W$, $H$, and $C$ are the width, the height, and the number of the feature maps, respectively. Suppose the kernel size $k$ of DW-Conv is 3, the number of convolutional groups is $g$, the theoretical complexity of the model under ideal conditions is
\begin{equation}
	\hspace{-1.5mm}
\begin{array}{l}
	\begin{aligned}
	\! F \! L \! O \! P \! s{_{2SConv}} \! &= \!\sum {\!2 \!\times \! W \! H \! \times \left( {\! k \!\times \! k  \!\times {\! C_{\! in}}/ \! g \! + \! 1} \right) \! \times {\! C_{\! out}}}\\
	 \! &= \left( {46 + 8C} \right) \! \times WHC
	\end{aligned}
\end{array}.
	\label{eq:1}
\end{equation}
From Eq.~\ref{eq:1}, the key variable that affects FLOPs is the quadratic term of $C$. As shown in Fig.~\ref{fig:2}, because the number of feature maps input to SConv Block is usually very large (the smallest one also reaches 32+128=160, $C \gg 5.75, 8C \gg 46$), the operation of generating the quadratic term of $C$, i.e., the 1×1 convolution, is one of the main factors affecting the computational complexity. This impact is more obvious in other branches with more feature maps. Besides, the authors \cite{ma2018shufflenet} suggested that the inference speed of the model may also be related to memory access cost and equal channel width minimizing this metric. The 1×1 convolutions in SConv Blocks change the number of feature maps frequently, which may also cause a bottleneck in model inference efficiency.

2) The imbalanced features

In CNN, the feature maps output by the shallow layer of the backbone network usually has low-level semantic features and high resolution, which is more helpful to detect small objects that lack detailed semantic features but requiring higher resolution for localization. While feature maps output from deep layers of the backbone network usually has lower resolution and contain rich abstract semantic features. A wise approach is to integrate these contextual features output by the CNN backbone to make their advantages complementary.

However, backbone networks, such as MobileNet \cite{howard2017mobilenets} and ShuffleNet \cite{zhang2018shufflenet}, are all designed for image classification tasks. Since the image classification task only needs the features output by the last layer, these lightweight networks usually have far fewer feature maps output from the shallow layers for compressing the model. Therefore, directly replacing the backbone of a lightweight detector with these backbone networks may lead to the feature imbalance problem at different levels.

From the first branch of the neck in Fig.~\ref{fig:2}, it can be seen that the number of shallow feature maps is 32, which is much smaller than the number of deep feature maps. In the feature fusion process, the low-level high-resolution features are submerged in the high-level low-resolution features, and this is very unfavorable for the detection of small objects. In addition, other feature imbalance issues, such as the lack of feature propagation paths from shallow layer to deep layer, and the unsuitable size of receptive fields, are also bottlenecks that restrict the improvement of the detection accuracy.

3) Unable to detect OBBs

General detectors are only suitable for detection tasks of HBBs, but for RSOD tasks, both HBBs and OBBs may exist. The baseline model cannot work when objects are represented by OBBs. For angle-based OBB detection heads \cite{fuRotationawareMultiscaleConvolutional2020}, anchor-based detectors usually need to set more anchor boxes with different rotation angles and different proportions in advance, which seriously affects the inference speed. In addition, it can only obtain rectangular bounding boxes. For the vertex-based OBB detection head \cite{xu2020gliding}, the problem that vertex regression is difficult to approach the boundary value, and the problem of many degrees of freedom of the vertices, are needed to be considered when designing an OBB detection head.

\section{Proposed LO-Det Method}
\begin{figure*}[htbp]
	\centering
	\epsfig{width=0.9\textwidth,file=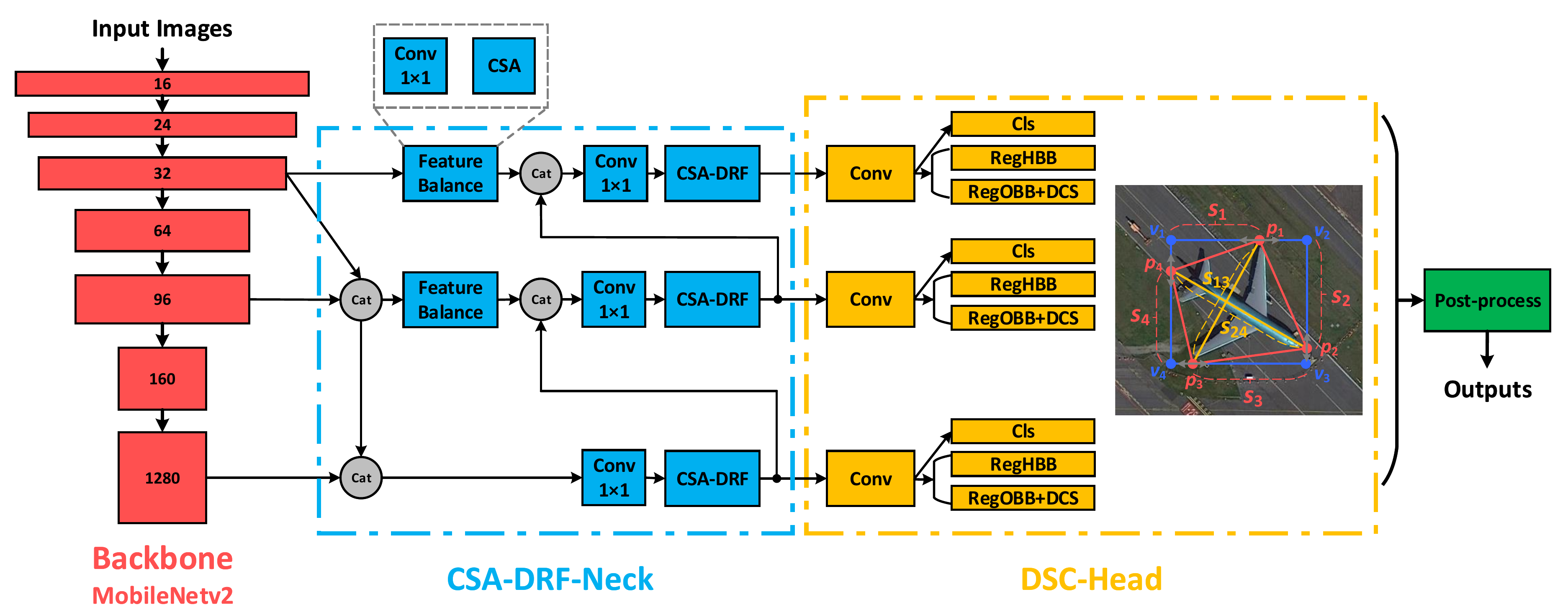}
	\caption{Structure of the proposed LO-Det including CSA-DRF module, feature balance module, and DSC-Head.}\label{fig:4}
\end{figure*}

In response to the problems of existing methods, an effective lightweight oriented object detector, LO-Det, is proposed. The model structure of the proposed LO-Det is shown in Fig.~\ref{fig:4}. It is mainly composed of three parts: backbone, neck with CSA and DRF blocks, and DSC-Head. Among them, the key ingredients of the proposed LO-Det are the neck component and the head component. This is because the neck and head components occupy 4.403 Giga floating-point operations (GFLOPs) (about 65\%) in the 6.753 GFLOPs of the baseline model. They are respectively designed for improving efficiency of the neck component in the existing lightweight detectors, and detecting OBBs in some RSOD tasks. Specifically, the neck component consists of CSA blocks with higher efficiency than stacked separable convolution, and DRF blocks for fusing context features of the network. The head component of the proposed LO-Det is a DSC-Head module improved from the gilding vertex method \cite{xu2020gliding}, which helps to predict OBBs more stably and accurately. Besides, the widely adopted MobileNetv2 \cite{sandler2018mobilenetv2} with good performance is selected as the backbone. This backbone only consists of separable convolutional layers and 1×1 vanilla convolutional layers, without other additional operations. The simplicity and high repeatability of the model's constituent elements facilitate further pruning, quantification, and CUDA optimization during actual deployment. In the subsequent experimental part, the performance of other state-of-the-art backbones is also compared and evaluated. 

\subsection{CSA-DRF Neck}

In view of the inefficiency of stacked separable convolutions and feature imbalance problems analyzed above, a novel neck component structure with CSA-DRF sub-modules is proposed.

1) Channel Separation-Aggregation (CSA) block

\begin{figure}[htbp]
	\centering
	\epsfig{width=0.49\textwidth,file=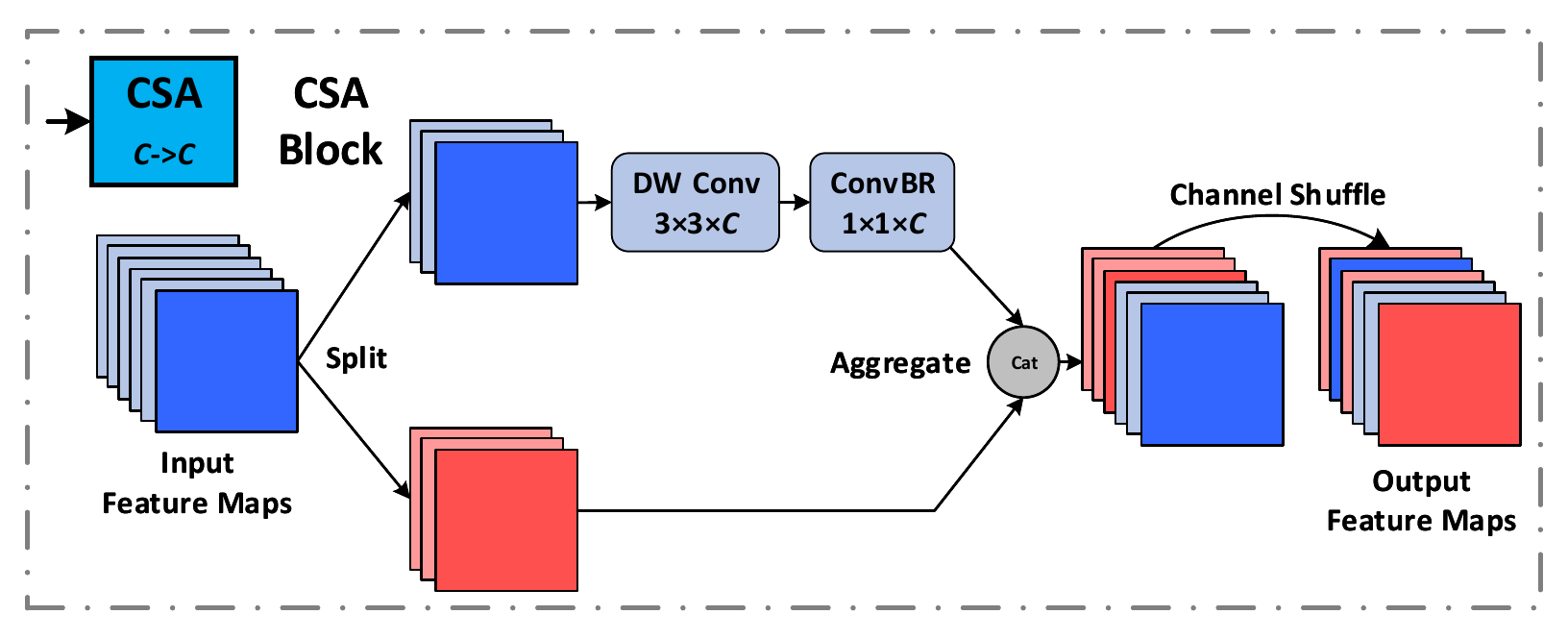}
	\caption{Structure of CSA block.}\label{fig:5}
\end{figure}

According to the above analysis, the 1×1 convolution operation in the neck component of a lightweight detector is the main consumer of inference time. Too many 1×1 convolutions and changing the input and output of a 1×1 convolution frequently may reduce the computational efficiency. The intuitive idea is to reduce the filters (channels) of 1×1 convolutional layers and make the input and output the same. GhostNet \cite{han2020ghostnet} observes that there are similarities and redundancy between features. Based on this observation, the CSA structure is designed as shown in Fig.~\ref{fig:5}. In CSA block, only half of the features are subjected to SConv operation, and then superimposed with the other half. This reduces the computation of 1×1 convolution and keeps the number of input and output feature maps consistent. However, because the features of different branches are isolated from each other and there is no information exchange, it causes performance degradation \cite{howard2017mobilenets}. Thus, the channel shuffle \cite{zhang2018shufflenet} operation is used to make feature interactions across branches.

After these improvements, it can be found that the structure of CSA block is similar to that of ShuffleNetv2 \cite{ma2018shufflenet}, although based on different starting points. The difference between them is that the CSA structure does not have a 1×1 convolution before the 3×3 DW-Conv and the number of feature maps is the same. There are two main reasons for doing so. First, based on the aforementioned analysis that changing the number of feature maps frequently leads to a decrease in efficiency, the number of features is designed to be unchanged as in CSA, so there is no need to use 1×1 convolution for feature alignment before DW-Conv. While in ShuffleNetv2 \cite{ma2018shufflenet}, the number of feature maps of the intermediate layer (DW-Conv) is inconsistent with the input and output, so 1×1 convolution is required for feature alignment. Secondly, each CSA block connected in series has a channel shuffle module at the end that can replace the 1×1 convolution at the forefront of the next CSA block for features exchange. 

\begin{figure}[htbp]
	\centering
	\epsfig{width=0.45\textwidth,file=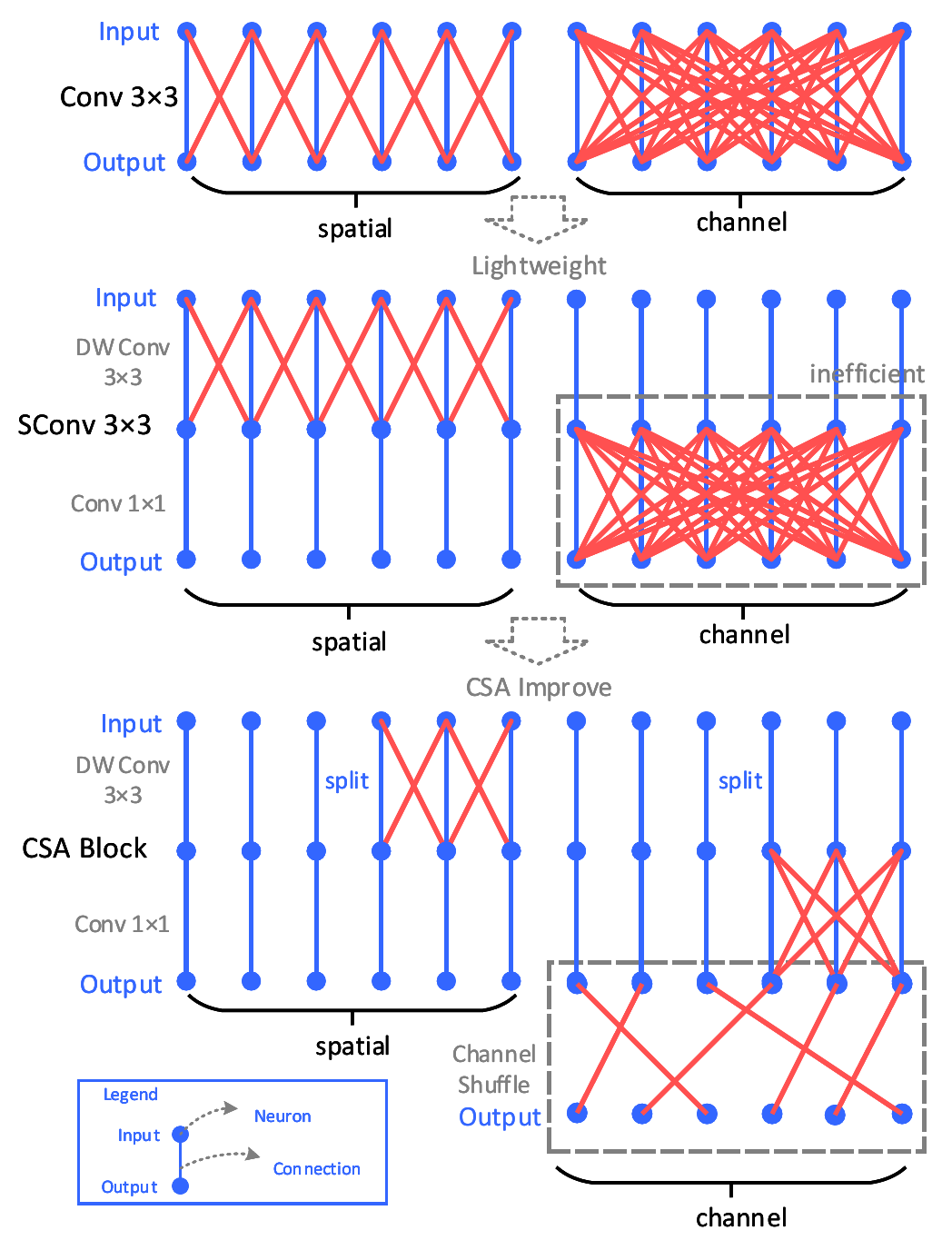}
	\caption{The principle and neuron connection relationship of the proposed CSA structure.}\label{fig:6}
\end{figure}

Fig.~\ref{fig:6} shows the principle and neuron connection relationship of the proposed CSA structure to achieve lightweight. Compared with the vanilla 3×3 convolution, SConv reduces the computational complexity by decomposing the space and channel operations by DW-Conv and 1×1 convolution. CSA further decomposes the features on channel direction and uses the cheaper operation, channel shuffle, for Cross-channel feature exchange. In order to make a fair comparison with SConv block, every two CSAs are connected as a group and calculate it FLOPs. The FLOPs of two CSA blocks is
\begin{equation}
	\hspace{-1.2mm}
	\begin{array}{l}
	\begin{aligned}
		F \! L \! O \! P \! s{_{2CSA}}
		 &= \left( {22 + C} \right) \! \times \! WHC \! + 2 \! F \! L \! O \! P \! s{_{\! shuffle}}.
	\end{aligned}
	\end{array}
	\label{eq:2}
\end{equation}
ShuffleNet [19] indicates that 
\begin{equation}
	F \! L \! O \! P \! s{_{shuffle}} \ll \! F \! L \! O \! P \! s{_{1 \times 1{\rm{ }}conv}} \! = 2\left( {C + 1} \right) \! \times WHC ,
	\label{eq:3}
\end{equation} 
so
\begin{equation}
	F \! L \! O \! P \! s{_{2 C \! S \! A}} \ll \left( {26 \! + 5C} \right) \! \times \! WHC \! < \! F \! L \! O \! P \! s{_{2SConv}}.
	\label{eq:4}
\end{equation}
Therefore, the proposed CSA structure has lower computational complexity compared with the original stacked SConvs.

2) Dynamic Receptive Field (DRF) block

Although the CSA structure reduces the computational complexity, too sparse connections in the neural network also reduce the feature extraction performance (the approximation ability of nonlinear models is reduced). It can be observed from Fig.~\ref{fig:6} that this problem is mainly due to the DW-Conv operation performed on only half of the feature maps. 

\begin{figure}[htbp]
	\centering
	\epsfig{width=0.49\textwidth,file=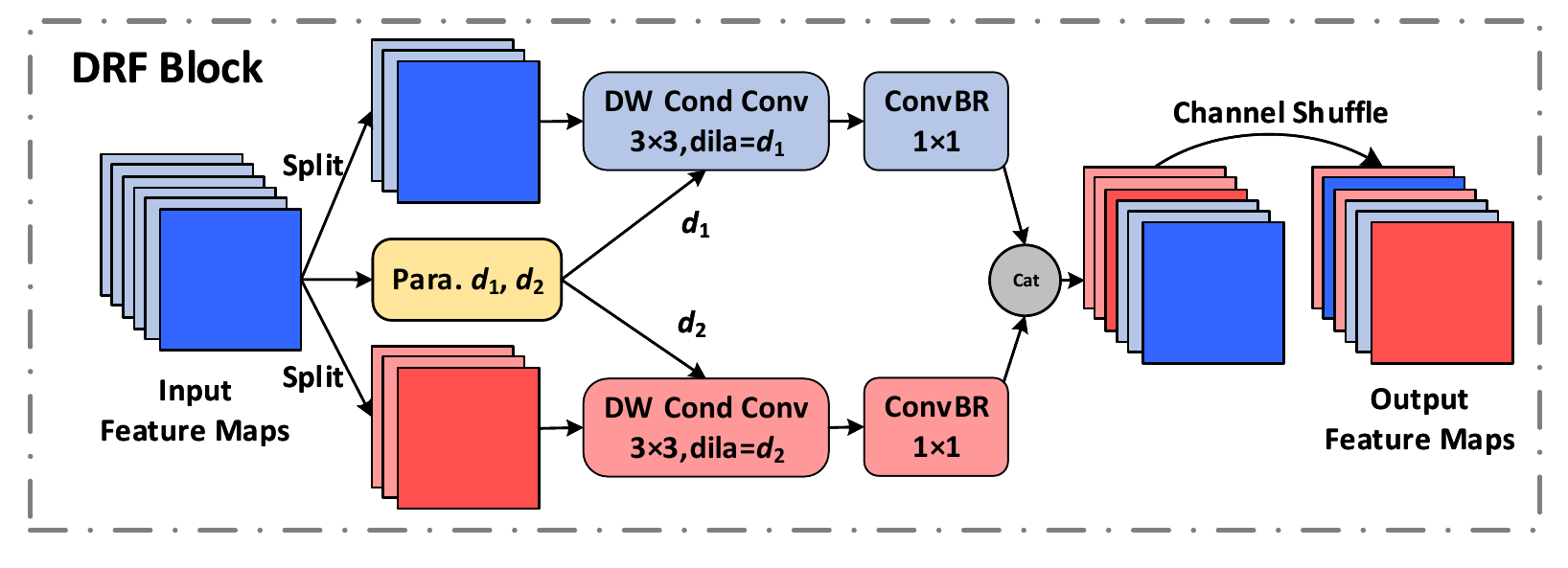}
	\caption{Structure of DRF block.}
	\label{fig:7}
\end{figure}

Therefore, the CSA structure is modified by using SConv in both branches to improve the feature extraction performance. Furthermore, considering to enhance feature interactions in the spatial direction (for example, the feature interactions of 3×3 DW-Conv in Fig.~\ref{fig:6} is only carried out in adjacent neurons) and the problem that receptive fields of the model may not be suitable for objects with any sizes, a dynamic receptive fields (DRF) structure is proposed as shown in Fig.~\ref{fig:7}. In DRF blocks, dilated convolution is used for a larger range of feature interactions in the spatial dimension, of which the dilated rate is designed to be learned through the network so that receptive fields of the model are adjusted dynamically to cover different objects. Also, the dilated rates can be set to static values through statistical analysis of the samples’ sizes, which saves some inference time. In order to further enhance spatial feature interactions, using a larger convolutional kernel is also an option, but it takes too much time. Inspired by the dynamic characteristics of neural networks, conditional convolution \cite{yang2019condconv} is introduced to improve 3×3 DW-Conv, in which a convolution tailored for the input is obtained using the idea of dynamic weighting for getting more efficient performance improvement than increasing the size of the convolution kernel. In dilated DW-Conv, different dilated rates are introduced to get wider feature interaction in the spatial dimension. In the channel dimension, there are more feature interactions and neuron connections than the CSA structure.

\begin{figure}[htbp]
	\centering
	\epsfig{width=0.49\textwidth,file=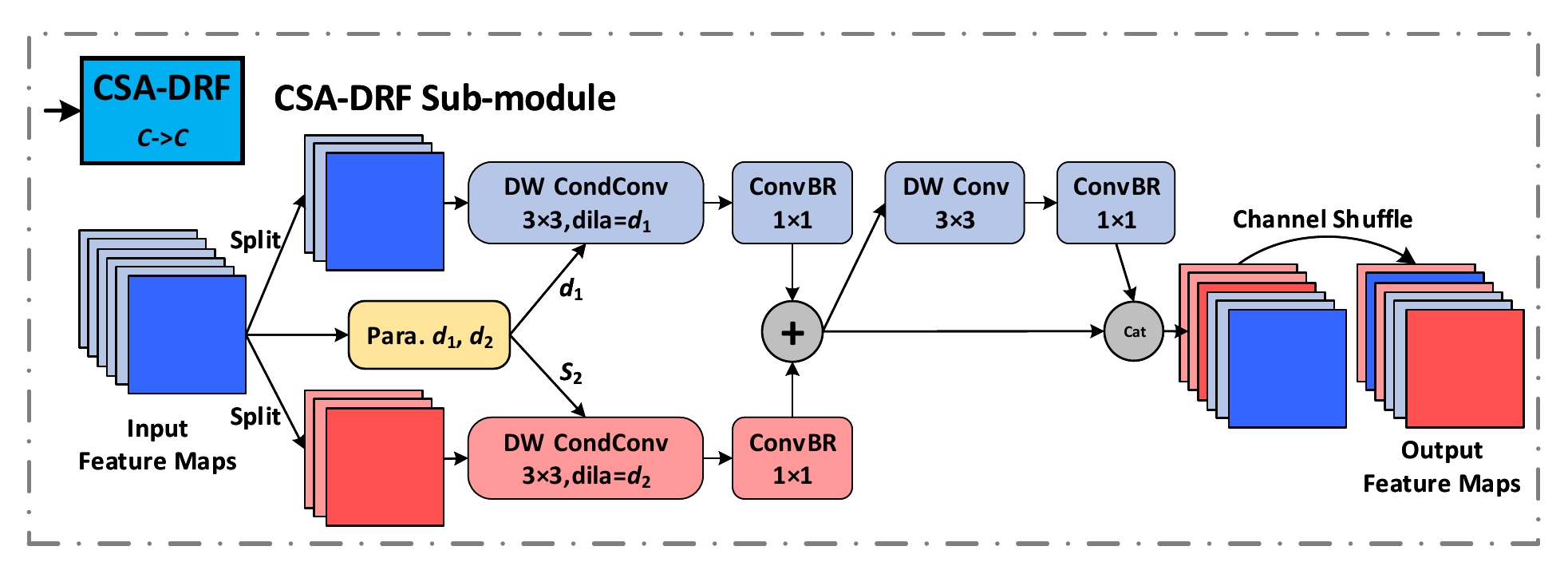}
	\caption{Structure of CSA-DRF sub-module.}
	\label{fig:9}
\end{figure}

\begin{figure}[htbp]
	\centering
	\epsfig{width=0.45\textwidth,file=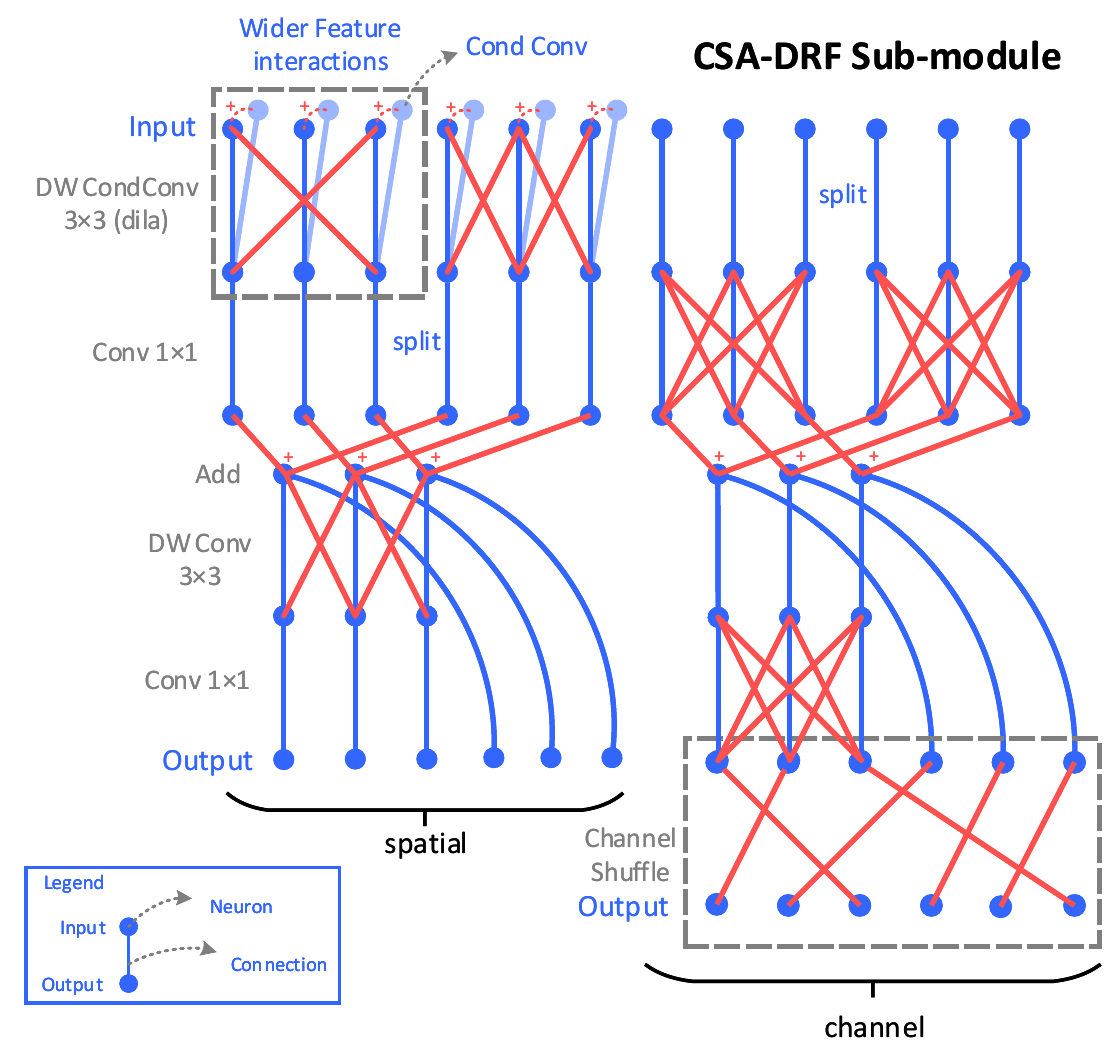}
	\caption{The principle and neuron connection relationship of the proposed CSA-DRF structure.}
	\label{fig:10}
\end{figure}

However, the DRF structure also brings higher computational complexity. Therefore, it is not recommended to replace each CSA structure with a DRF structure, but only one of the two-by-two CSA blocks by DRF structure. Besides, the time cost of channel shuffle cannot be ignored. From this perspective, the new CSA-DRF sub-module is obtained by fusing the CSA and DRF structures as shown in Fig.~\ref{fig:9}, and the principle and neuron connection are shown in Fig.~\ref{fig:10}. In the CSA-DRF sub-module, the first channel shuffle is removed and addition operation is used instead. This is because both branches have performed 1×1 convolution, and the feature interaction between channels is more sufficient than that of the CSA block. And if the concatenation operation is still used without channel shuffling, the features of the two branches are still independent of each other without interactions in the next CSA block. Finally, in the new neck component, a CSA-DRF sub-module is used to replace two original SConv blocks or two CSA blocks.

The complexity of the proposed CSA-DRF sub-module is
\begin{equation}
\hspace{-1mm}
\begin{array}{l}
	\begin{aligned}	
	\! F \! L \! O \! P \! s{_{C \! S \! A{\rm{ \!- }}\! D \! R \! F}} 
	 \!&= \left( {33 \! + {\textstyle{{3C} \over 2}}} \right) \! \times \! W \! H \! C \! + \! \! F \! L \! O \! P \! s{_{shuffle}},
	\end{aligned}
\end{array}
	\label{eq:5}
\end{equation} 
which is smaller than $FLOP{s_{2SConv}}$. This means that the proposed CSA-DRF sub-module has lower computational complexity and better efficiency than the SConv blocks in the baseline model.

3) Feature balance mechanism
 
In the aforementioned analysis, it is pointed out that because the most of lightweight backbone networks are designed for image classification tasks, the output features from their shallow layers and deep layers have obvious differences in quantity. This problem leads to a small amount of high-resolution low-level features being submerged in a large number of low-resolution high-level semantic features, which is not beneficial to RSOD tasks that need to utilize these features comprehensively.

Therefore, a feature balancing sub-module composed of 1×1 convolution and CSA blocks is designed and introduced into the neck component. 1×1 convolution is used to adjust the number of feature maps, and the CSA module performs feature fusion. Many studies \cite{tanEfficientDetScalableEfficient2020} have demonstrated that the CNN models with balanced depth (the number of convolutional layers), width (the number of feature maps in each layer, that is, $C$), and the size of feature maps ($W$ and $H$) usually achieve better performance. Therefore, this work also follows this principle: when $W$ and $H$ are reduced by half, the number of feature maps doubles. The complexity of feature balance sub-module is
\begin{equation}
	\begin{array}{l}
		\begin{aligned}	
	\! F \! L \! O \! P \! s{_{\! F \! B}}
	 \!&= \left( {1{\rm{3}} \! + 2{C_{in}}{\rm{ + }}{\textstyle{C \over 2}}} \right) \! \times \! W \! H \! C + \! F \! L \! O \! P \! s{_{\! shuffle}},
		\end{aligned}
	\end{array}
	\label{eq:6}
\end{equation}
where ${C_{in}}$ is the number of feature maps input from the shallow layers of network. Besides, down-sampling is also added at the connection between the backbone and neck components to further promote the propagation of features from the shallow layers of the network to the deep layers of the network.

In summary, the complexity of a branch (including feature balance sub-module) of the proposed CSA-DRF component is
\begin{equation}
	\begin{array}{l}
		\begin{aligned}	
\! F \! L \! O \! P \! s{_{newneck}} 
 &= \left( {{\rm{48 + 6}}C + {\rm{2}}{C_{in}}} \right) \! \times W \! H \! C \\ &+ 2 \! F \! L \! O \! P \! s {_{shuffle}}\\
 &< \left( {{\rm{50 + 10}}C + {\rm{2}}{C_{in}}} \right) \! \times \! W \! H \! C
		\end{aligned}
	\end{array}.
	\label{eq:7}
\end{equation}
The complexity of a branch of the baseline neck component is 
\begin{equation}
	\begin{array}{l}
		\begin{aligned}	
\! F \! L \! O \! P \! s{_{baseneck}} \! & = \left( {{\rm{94}} + {\rm{16}}C{\rm{ + 2}}{C_{in}}} \right) \! \times \! W \! H \! C
		\end{aligned}
	\end{array}.
	\label{eq:8}
\end{equation}
From above analysis, the complexity of the proposed CSA-DRF component is lower than that of the baseline neck component. In addition, it should be noted that the above case is only an ideal value for preliminary estimation, and some other situations have not been considered. For example, the last branch does not have a feature balance sub-module, and the complexity of some other operations, such as up-sampling, has not been calculated.

\subsection{DSC-Head}

In some remote sensing scenarios, OBBs can better describe the boundaries and shapes of the objects than HBBs. However, general detectors cannot detect OBBs, and there is no lightweight detector for detecting OBBs in the field of remote sensing. Since the gliding vertex method \cite{xu2020gliding} is easily extended based on HBBs and does not add too much complexity, it is employed into the proposed lightweight detection head component. But this method also has the two problems that need to be dealt.

1) Control the shapes of OBBs

\begin{figure}[htbp]
	\centering
	\epsfig{width=0.49\textwidth,file=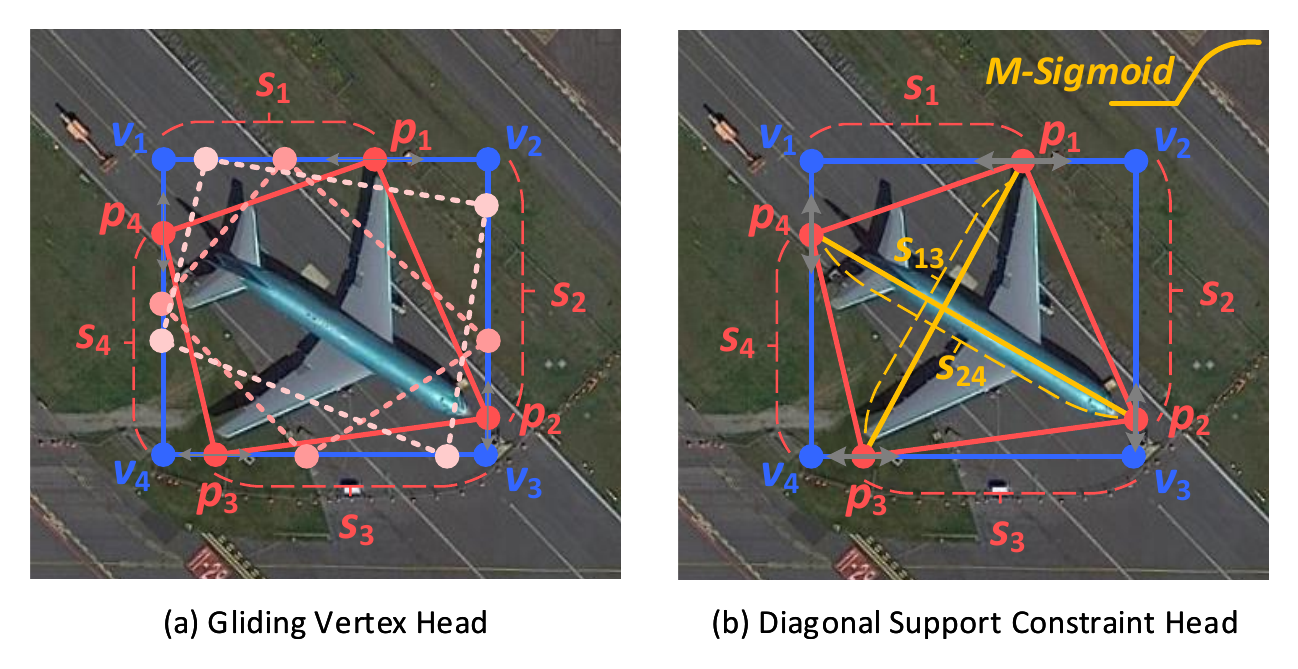}
	\caption{The principle of the proposed DSC-Head.}
	\label{fig:11}
\end{figure}

\begin{figure}[htbp]
	\centering
	\epsfig{width=0.49\textwidth,file=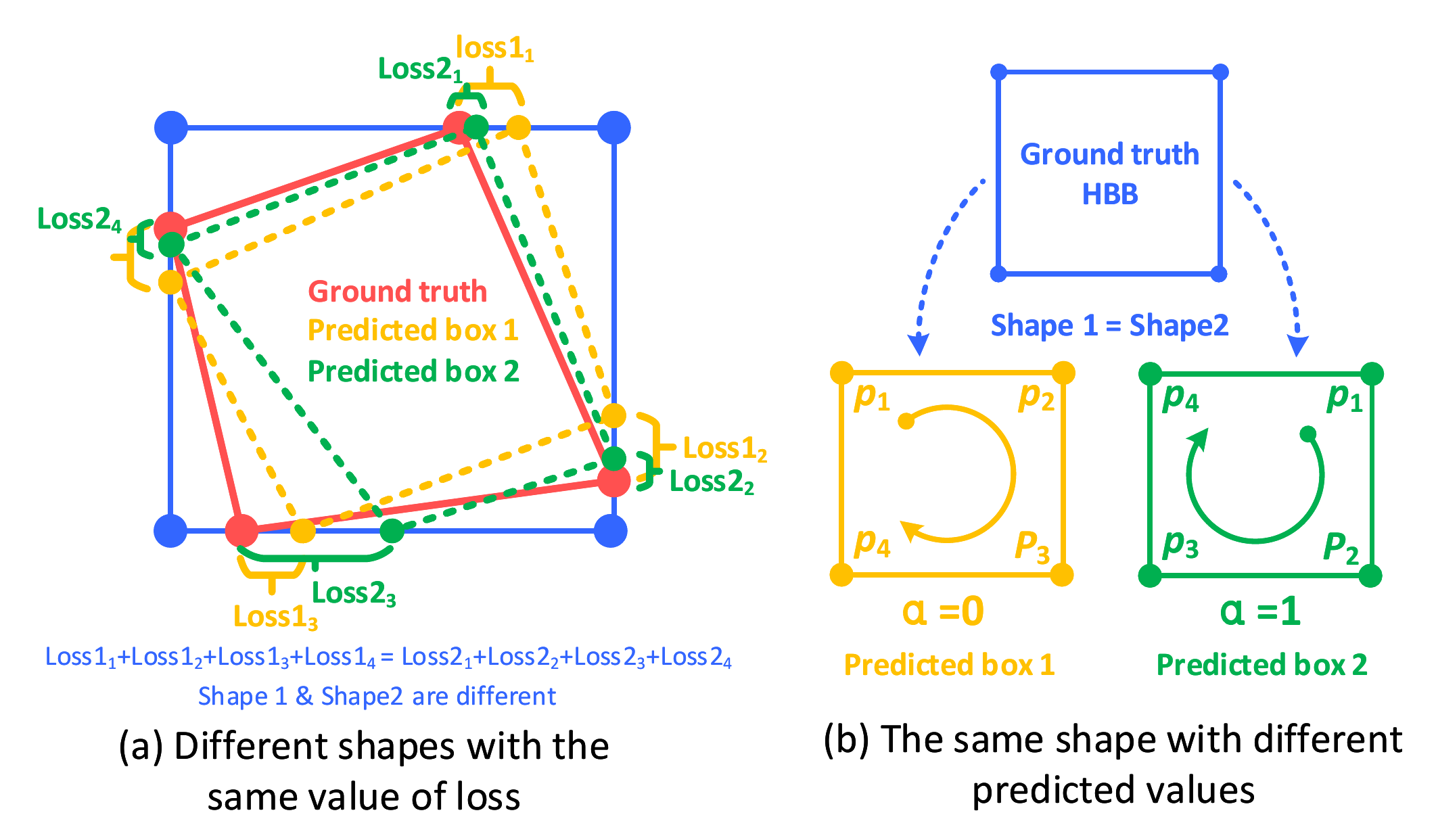}
	\caption{Analysis of the problems in vertex-based methods. (a) The problem of different shapes with the same value of loss. (b) The problem of the same shape with different predicted values.}
	\label{fig:12}
\end{figure}

As shown in Fig.~\ref{fig:11}, because the gliding vertices have too much freedom, it is difficult to control the shapes of OBBs regressed by the neural network. For example, as shown in Fig.~\ref{fig:12} (a), consider two cases, one is that the position loss of four vertices is equal, and the other is that the position loss of three vertices is small and the position loss of the other vertex is large. Although the final loss in the two cases may be the same, the shapes of the OBBs obtained is completely different. Since OBBs have more diverse shapes than HBBs, this makes the predictions more uncertain and may affect the detection accuracy. 

In this regard, a detection head component with diagonal support constraints is proposed, and the shapes of OBBs are controlled by constraining the diagonal length ratio of the quadrilateral. As shown in Fig.~\ref{fig:11}, suppose the vertices of an HBB are ${v_1},{v_2},{v_3},{v_4}$; the vertices of an OBB are ${p_1},{p_2},{p_3},{p_4}$; the center point of the HBB is $(x,y)$; the width of the HBB is $w$; the height of the HBB is $h$; the gliding distances of ${p_1},{p_2},{p_3},{p_4}$ are ${s_1},{s_2},{s_3},{s_4}$, respectively; and the diagonal length are ${s_{13}}$ and ${s_{24}}$. Therefore, the HBB’s variables predicted by the end-to-end CNN detector are
\begin{equation}
	\begin{array}{l}
		\begin{aligned}	
{t_x} \! = \frac{{x - {x_a}}}{{{w_a}}},{\rm{ }}{t_y} \! = \frac{{y - {y_a}}}{{{h_a}}},{\rm{ }}{t_w} \! = \log \frac{w}{{{w_a}}},{\rm{ }}{t_h} \! = \log \frac{h}{{{h_a}}}
		\end{aligned}
	\end{array}.
	\label{eq:9}
\end{equation}
where $({x_a},{y_a})$, ${w_a}$, and ${h_a}$ represent the center point, width and height of the anchor box, respectively; ${t_x},{t_y},{t_w},{t_h}$ are the variables directly predicted by the neural network. The OBB’s variables predicted by the end-to-end CNN detector are
\begin{equation}
	\begin{array}{l}
		\begin{aligned}	
\left\{ \begin{array}{l}
	{\alpha _1} = \frac{{{s_1}}}{w},{\rm{ }}{\alpha _2} = \frac{{{s_2}}}{h},{\rm{ }}{\alpha _3} = \frac{{{s_3}}}{w},{\rm{ }}{\alpha _4} = \frac{{{s_4}}}{h}\\
	{\beta _1} = \frac{{{s_{13}}}}{w},{\beta _2} = \frac{{{s_{24}}}}{h}
\end{array} \right.
		\end{aligned}
	\end{array}.
	\label{eq:10}
\end{equation}
In this way, OBB can be calculated according to the variables obtained from the above prediction based on HBB. In fact, because DSC-Head adds ${s_{13}}$ and ${s_{24}}$ constraints, ${\alpha _3}$ and ${\alpha _4}$ do not need to be predicted, that is, an OBB is determined by predicting ${\alpha _1},{\alpha _2},{\beta _1},{\beta _2}$.

2) Boundary value

Because the neural network has a large regression range for the variables ${\alpha _1},{\alpha _2},{\beta _1},{\beta _2}$ that may lead to unstable predictions, the Sigmoid function is used to limit the regression range. But for a Sigmoid function, it is difficult to regress to the boundary value of 0 or 1 (the bounding box is HBB in this case) accurately. In this regard, the H-Sigmoid function 
\begin{equation}
H{\rm{ - }}Sigmoid{\rm{ = }}\min \left( {\max \left( {0,x + 3} \right),6} \right)/6,
	\label{eq:11}
\end{equation}
where hard boundary values are introduced. 

\begin{figure}[htbp]
	\centering
	\epsfig{width=0.49\textwidth,file=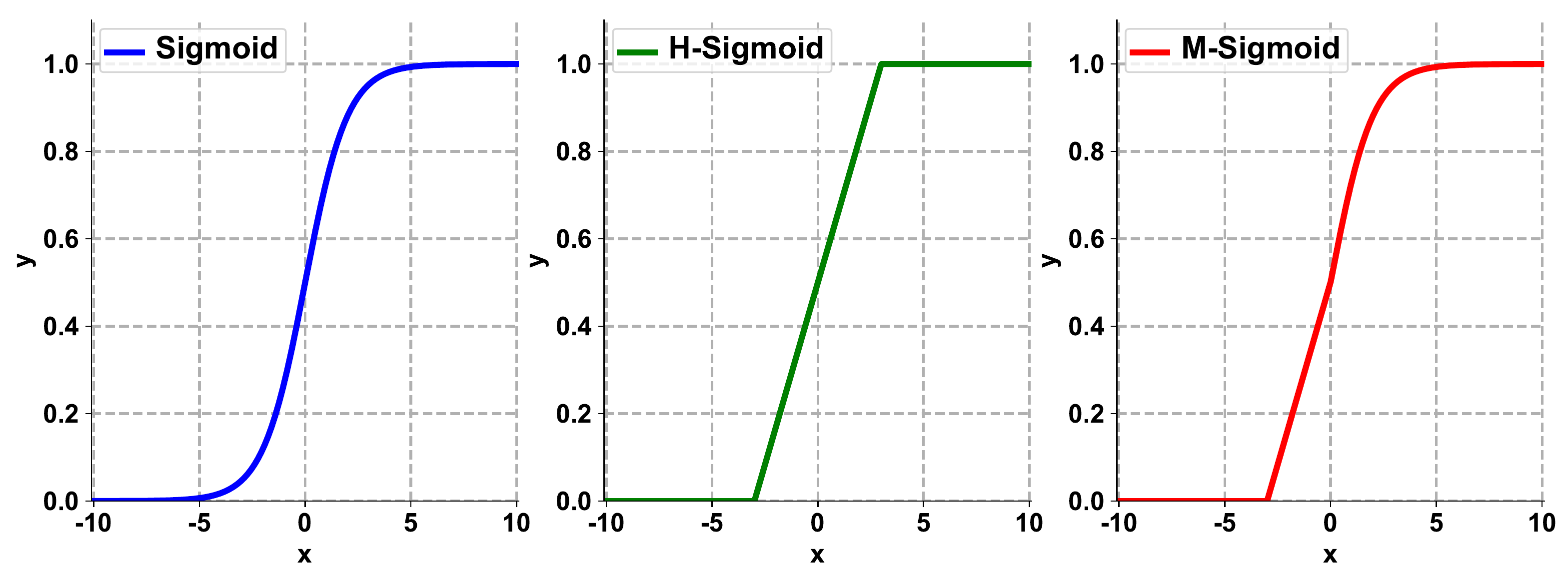}
	\caption{The proposed M-Sigmoid function. It should be explained that because the original piecewise function M-Sigmoid is difficult to derive in the neural network, such an approximation scheme: \textit{(clamp(sigmoid(x),0.01,1)-0.01)/(1-0.01)} is used in the program.}
	\label{fig:13}
\end{figure}

However, it is also necessary to consider these two cases, in which ${\alpha_1},{\alpha_2},{\beta_1},{\beta_2}$ are all 0 or 1, as shown in Fig.~\ref{fig:12} (b). They both represent the same HBB but the predicted values are different, which is ambiguous. Therefore, as shown in Fig.~\ref{fig:13}, the half of H-sigmoid is replaced by a Sigmoid function with soft boundary values to obtain a mixed Sigmoid (M-Sigmoid) function, 
\begin{equation}
M{\rm{-}}Sigmoid{\rm{\!=}}\left\{ \begin{array}{l}
	\! \min \left( {\max \left( {0,x + 3} \right),6} \right)/6{\rm{  }},  \rm{  }x \le 0\\
	\! 1/\left( {1 + {e^{ - x}}} \right){\rm{    }},  \rm{  }otherwise
\end{array} \right.
	\label{eq:12}.
\end{equation}
Thus, 
\begin{equation}
{\alpha _i} = M{\rm{\!-}}Sigmoid\left( {{t_{{\alpha _i}}}} \right),{\rm{  }}{\beta _i} = M{\rm{ - }}Sigmoid\left( {{t_{{\beta _i}}}} \right),
	\label{eq:13}
\end{equation}
where the OBB variables ${\alpha _i},{\beta _i} \in \left[ {{\rm{0,1}}} \right)$, ${t_{{\alpha _i}}},{t_{{\beta _i}}}$ are predicted by the network. It should be explained that because the original piecewise function M-Sigmoid is difficult to derive in the neural network, such an approximation scheme: \textit{(clamp(sigmoid(x),0.01,1)-0.01)/(1-0.01)} is used in the program. It solves the problem that the boundary value is difficult to regress and alleviates the ambiguity of OBB representation.

3) Loss function

The loss function of DSC-Head consists of four parts: confidence loss $Los{s_{conf}}$, HBB loss $Los{s_{HBB}}$, OBB loss $Los{s_{OBB}}$, and classification loss $Los{s_{cls}}$,
\begin{equation}
Loss = Los{s_{conf}} + Los{s_{HBB}} + Los{s_{OBB}} + Los{s_{cls}},
	\label{eq:14}
\end{equation}

\begin{equation}
\begin{array}{l}
	\begin{aligned}	
	Los{s_{OBB}} &= \sum\limits_{i = 1}^4 {smooth{\rm{ - }}L1\left( {{\alpha _i} - \alpha _i^{gt}} \right)} \\
	& + \sum\limits_{j = 1}^2 {smooth{\rm{ - }}L1\left( {{\beta _j} - \beta _j^{gt}} \right)} 
	\end{aligned}
\end{array},
\label{eq:15}
\end{equation}
where $\alpha _i^{gt}$ and $\beta _j^{gt}$ denote the ground truth values of ${\alpha _i}$ and ${\beta _j}$. The loss of other parts is the same as the baseline.

\section{Experiments and Discussions}
In this section, experiments on public remote sensing datasets are conducted to verify the effectiveness of the proposed LO-Det and further evaluate its detection performance. First, the experimental conditions are explained. Secondly, the ablation experiments are conducted and the performance of each improvement is discussed. Third, the influence of some important parameters on model performance is evaluated. Finally, comparative experiments on large public datasets are provided to compare and evaluate the performance of the proposed LO-Det with the state-of-the-art methods.

\subsection{Experimental Conditions}

1) Experimental platforms

\begin{figure}[htbp]
	\centering
	\epsfig{width=0.45\textwidth,file=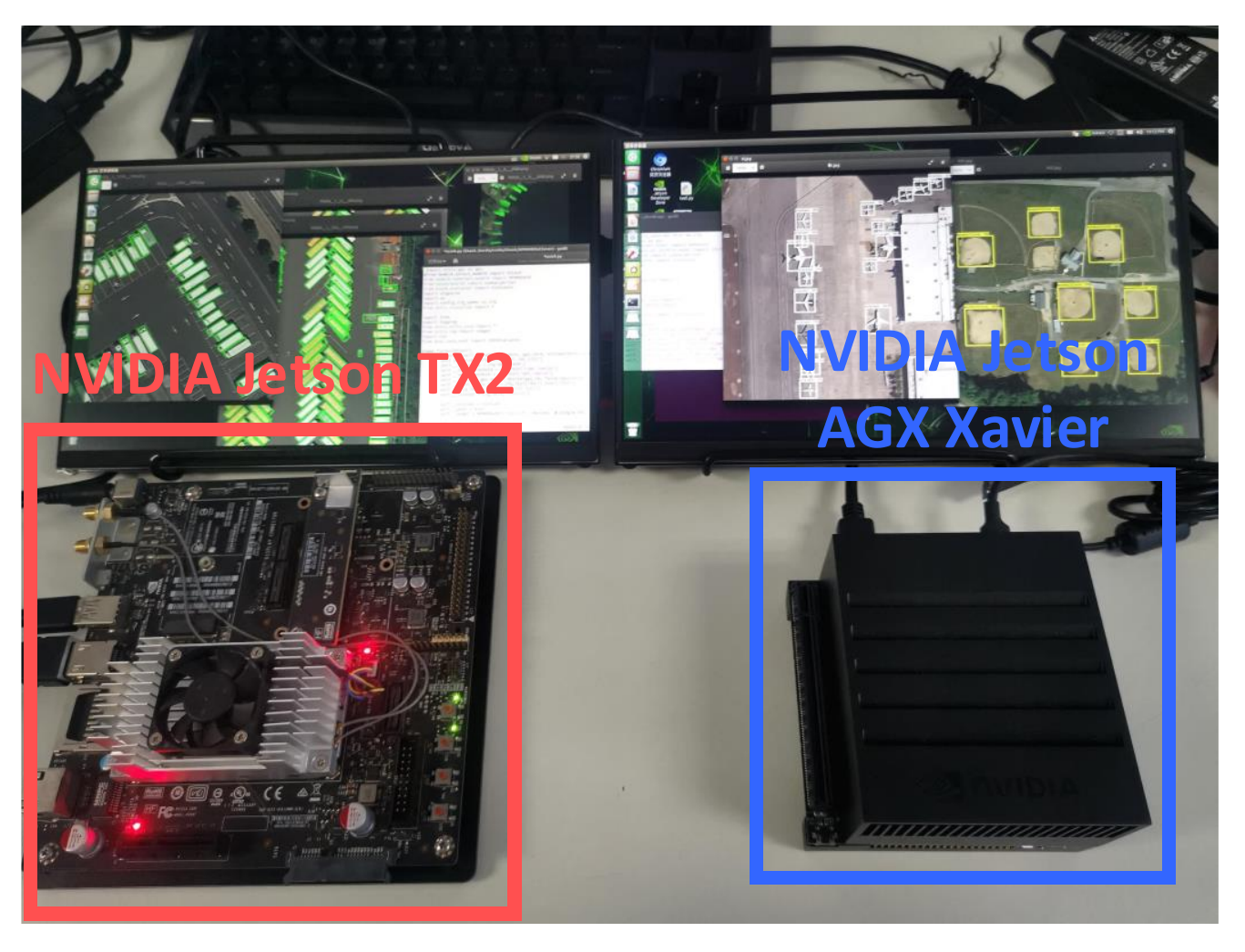}
	\caption{The experimental embedded devices.}
	\label{fig:14}
\end{figure}

The proposed LO-Det is expected to run not only on expensive GPUs, such as RTX3090, but also ordinary GPUs, such as GTX 1660 that most people can afford, and even on embedded devices. Therefore, in order to evaluate the performance of the proposed LO-Det comprehensively and provide baselines for this new direction, a variety of experimental platforms are used, including: a) a computer with an Intel Core i7-10700K CPU (3.80GHz), 64 GB of memory, and an NVIDIA GeForce RTX 3090 GPU (24GB); b) a computer with an Intel Core i7-8700 CPU (3.20GHz), 16 GB of memory, and an NVIDIA GeForce GTX 1660 GPU (6GB); c) an NVIDIA Jetson TX2 embedded device; d) an NVIDIA Jetson AGX Xavier embedded device. The embedded devices are shown in Fig.~\ref{fig:14}.

2) Datasets

DOTA \cite{xiaDOTALargeScaleDataset2018} and DIOR \cite{liObjectDetectionOptical2020}, the largest and widely used high-resolution aerial remote sensing datasets in RSOD tasks, are used to verify and evaluate the proposed method. The selected datasets cover more than 20 categories of objects, and cover a variety of scenes. Such a wealth of experimental datasets help to verify and evaluate the performance of the proposed method more comprehensively.

a) DOTA dataset \cite{xiaDOTALargeScaleDataset2018} is a large aerial image dataset containing 2806 aerial images from 800 × 800 pixels to 4000 × 4000 pixels, in which more than 188,000 objects are annotated. These objects fall into 15 categories: Plane (PL), Baseball diamond (BD), Bridge (BR), Ground field track (GFT), Small vehicle (SV), Large vehicle (SV), Ship (SH), Tennis court (TC), Basketball court (BC), Storage tank (ST), Soccer-ball field (SBF), Roundabout (RA), Harbor (HA), Swimming pool (SP), and Helicopter (HC). Due to the huge size, these images are cropped into sub-images of 800×800 pixels with an overlap of 150 pixels.

b) DIOR \cite{liObjectDetectionOptical2020} is another large aerial image dataset with 23,463 images (800 × 800 pixels) containing 190,288 objects belonging to 20 categories: Airplane (c1), Airport (c2), Baseball field (c3), Basketball court (c4), Bridge (c5), Chimney (c6), Dam (c7), Expressway service area (c8), Expressway toll station (c9), Golf course (c10), Ground track field (c11), Harbor (c12), Overpass (c13), Ship (c14), Stadium (c15), Storage tank (c16), Tennis court (c17), Train station (c18), Vehicle (c19), and Windmill (c20).

3) Evaluation metrics

The mean Average Precision (mAP), the widely used metric in OD tasks, is adopted for evaluating the detection accuracy. According to the widely recognized PASCAL VOC standard, the IoU threshold of mAP is 0.5. The inference time of detecting an image and the detected frames per second (fps) are used to evaluate the detection speed. The FLOPs is used to evaluate the complexity of a CNN model. And the memory space occupied by parameters of a model is used to evaluate the model size.

4) Implementation details

To compare the proposed LO-Det with state-of-the-art methods fairly, training hyperparameters are set to be the same as the methods compared. The initial learning rate is $1.5 \times 10^{-4}$, the final learning rate is $1 \times 10^{-6}$, and the learning rate is updated by cosine strategy. The maximum training epoch is 100. And the NMS threshold is 0.45. The initial weight of the backbone is pre-trained on the ImageNet dataset.

5) Comparative methods

Due to the lack of benchmarking methods for comparative experiments, some mainstream detectors are chosen for comparisons. For example, methods widely used in common object detection tasks like Faster R-CNN \cite{renFasterRCNNRealTime2017a}, SSD \cite{liuSSDSingleShot2016a}, and RetinaNet \cite{linFocalLossDense2017}, and methods designed for RSOD tasks like R2CNN \cite{jiang2017r2cnn}, SCRDet \cite{yangSCRDetMoreRobust2019}, and CAD-Net \cite{zhangCADNetContextAwareDetection2019} are chosen. It should be noted that, since these detectors are of non-lightweight design with high complexity and large models, the detection accuracy of some of them is higher than the proposed LO-Det that has much lower computational complexity.

\subsection{Ablation Experiments and Discussions}

1) Ablation experiments

Since the baseline method cannot detect OBBs, and the benchmarks for detecting HBBs and OBBs cannot be directly compared, the two groups of ablation experiments are conducted separately. Among them, the experiments of HBBs are performed on the DIOR dataset which is the existing largest HBB datasets for RSOD tasks. The experiments of OBBs are performed on the DOTA dataset, in which the objects are annotated using OBBs. The new baseline on DOTA dataset is the original baseline + gliding vertex \cite{xu2020gliding} head (GV-Head). The experiment separately tests and evaluates the performance of each improved module, and the experimental results are listed in Table~\ref{table:1} and Table~\ref{table:2}.

\begin{table*}[htbp]
	\centering
	\renewcommand\arraystretch{2}
	\setlength{\tabcolsep}{0.5mm}{
		\caption{\label{table:1}
			{Ablation experiments and evaluations of the proposed LO-Det on the DIOR dataset}}
		\resizebox{\textwidth}{!}{
		\begin{tabular}{c|ccc|cccccccc}
			\hline\hline
			\multirowcell{2}{Modules} & \multirowcell{2}{Baseline} & \multirowcell{2}{CSA-DRF} & \multirowcell{2}{Feature \\
			Balance} & \multirowcell{2}{mAP \\ (\%)} & \multirowcell{2}{Speed 1 \\ {(fps)}} & \multirowcell{2}{Speed 2 \\ {(fps)}} & \multirowcell{2}{Speed 3 \\ {(fps)}} & \multirowcell{2}{Speed 4 \\ {(fps)}} & \multirowcell{2}{FLOPs} & \multirowcell{2}{Parameters \\ {(MB)}} & \multirowcell{2}{Model Size \\ {(MB)}} \\  & & & & & & & & & &  &\\ \hline
			\multirowcell{4}{Selected\\ Module(s)} & $\checkmark$ & & & 56.73 & 59.81 & 28.71 & 6.37 & 20.77 & 6.753G & 8.55 & 33.20 \\
			& $\checkmark$ & $\checkmark$ & & 62.90 \scriptsize{(+6.17)}& 64.77 \scriptsize{(+4.96)} & 34.41 \scriptsize{(+5.70)} & 10.31 \scriptsize{(+3.94)} & 26.09 \scriptsize{(+5.32)} &  5.335G \scriptsize{(-1.418G)} & 6.21 \scriptsize{(-2.34)} & 24.29 \scriptsize{(-8.91)} \\
			& $\checkmark$ & $\checkmark$ & $\checkmark$ & 65.85 \scriptsize(+9.12)& 60.53 \scriptsize(+0.72) & 30.46 \scriptsize(+1.75) & 7.14 \scriptsize(+0.77) & 22.75 \scriptsize(+1.98) & 6.385G \scriptsize(-0.368G) & 6.90 \scriptsize(-1.65) & 26.95 \scriptsize(-6.25) \\
			\hline\hline
	\end{tabular}}
	\begin{tablenotes}
	\item {Note: Speed 1 is the speed on RTX 3090 GPU, speed 2 is the speed on RTX 1660 GPU, speed 3 is the speed on NVIDIA Jetson TX2, Speed 4 is the speed on NVIDIA Jetson AGX Xavier.}
	\end{tablenotes}
}
\end{table*}

\begin{table*}[htbp]
	\centering
	\renewcommand\arraystretch{2}
	\setlength{\tabcolsep}{0.5mm}{
		\caption{\label{table:2}
			{Ablation experiments and evaluations of the proposed LO-Det on the DOTA dataset}}
		\resizebox{\textwidth}{!}{
			\begin{tabular}{c|ccccc|cccccccc}
				\hline\hline
				\multirowcell{2}{Modules} & \multirowcell{2}{Baseline} & \multirowcell{2}{CSA-DRF} & \multirowcell{2}{Feature \\ Balance} & \multirowcell{2}{GV-Head} & \multirowcell{2}{DCS-Head}
			 & \multirowcell{2}{mAP \\ (\%)} & \multirowcell{2}{Speed 1 \\ {(fps)}} & \multirowcell{2}{Speed 2 \\ {(fps)}} & \multirowcell{2}{Speed 3 \\ {(fps)}} & \multirowcell{2}{Speed 4 \\ {(fps)}} & \multirowcell{2}{FLOPs} & \multirowcell{2}{Parameters \\ {(MB)}} & \multirowcell{2}{Model Size \\ {(MB)}} \\ & & & & & & & & & & & & &\\ \hline
				\multirowcell{4}{Selected\\ Module(s)} & $\checkmark$ & & &$\checkmark$& & 61.91 & 58.16 & 28.22 & 6.43 & 21.08 & 6.753G & 8.55 & 33.20 \\
				& $\checkmark$ & $\checkmark$ & &$\checkmark$& & 62.86 \scriptsize{(+0.95)} & 63.27  \scriptsize{(+5.11)} & 33.97 \scriptsize{(+5.75)} & 10.09 \scriptsize{(+3.66)} & 25.81 \scriptsize{(+4.73)} &  5.374G \scriptsize{(-1.379G)} & 6.24 \scriptsize{(-2.31)} & 24.29 \scriptsize{(-8.91)} \\
				& $\checkmark$ & $\checkmark$ & $\checkmark$ & $\checkmark$ &  & 64.03  \scriptsize{(+2.12)} & 60.03 \scriptsize{(+1.87)} & 30.14 \scriptsize{(+1.92)} & 7.23  \scriptsize{(+0.80)} & 23.20 \scriptsize{(+1.98)} &  6.424G \scriptsize{(-0.329G)} & 6.93 \scriptsize{(-1.62)} & 26.95 \scriptsize{(-6.25)} \\
								& $\checkmark$ & $\checkmark$ & $\checkmark$ & & $\checkmark$ & 66.17 \scriptsize{(+4.26)} & 60.03 \scriptsize{(+1.87)} & 30.13 \scriptsize{(+1.91)} & 7.25  \scriptsize{(+0.82)} & 23.20 \scriptsize{(+1.98)} &  6.424G \scriptsize{(-0.329G)} & 6.93 \scriptsize{(-1.62)} & 26.95 \scriptsize{(-6.25)} \\
				\hline\hline
		\end{tabular}}
		\begin{tablenotes}
			\item {Note: Speed 1 is the speed on RTX 3090 GPU, speed 2 is the speed on RTX 1660 GPU, speed 3 is the speed on NVIDIA Jetson TX2, Speed 4 is the speed on NVIDIA Jetson AGX Xavier.}
		\end{tablenotes}
	}
\end{table*}

It can be observed from Table~\ref{table:1} that the proposed CSA-DRF not only increases the mAP by 6.17 (+10.88\%) on the DIOR dataset, but also reduces the model complexity by 1.418 GFLOPs, which makes the inference speed significantly improved on different devices. This means that the FLOPs of the newly proposed neck and head components (4.035G) are reduced by 8.36\% compared to the FLOPs of the neck and head components of the baseline (4.403G) when the mAP is increased by 9.12\%. Furthermore, the number of parameters and the occupied memory space are smaller. On this basis, although the introduction of the feature balance sub-module has increased some computational complexity and parameters, the mAP of object detection has also increased by 16.08\% to 65.85. Finally, compared with the baseline model, the proposed LO-Det has faster detection speed, lower complexity, and fewer parameters. In addition, the detection accuracy is also improved in the proposed LO-Det. 

The experimental results on the DOTA \cite{xiaDOTALargeScaleDataset2018} dataset also reflect the same results. Moreover, when the proposed DSC-Head component is used to detect OBBs, compared with the Gliding Vertex method \cite{xu2020gliding}, the mAP of 2.14 is further improved without introducing additional complexity or reducing the detection speed. The ablation experiments on the DOTA dataset also demonstrate that the indicators of the proposed LO-Det are better than that of the baseline + gliding vertex model, i.e., it is faster and better on RSOD tasks.

More detailed results of the ablation experiments are given in Table~\ref{table:3} and Table~\ref{table:4}, which provide the APs of different categories on DIOR dataset and DOTA dataset.

\begin{table*}[htbp]
	\centering
	\renewcommand\arraystretch{1.5}
	\caption{\label{table:3}
		{More detailed mAP (\%) results of ablation experiments on the DIOR dataset}}
	\begin{threeparttable}
		\centering
		\resizebox{\textwidth}{!}{\setlength{\tabcolsep}{0.53mm}{
		\begin{tabularx}{\linewidth}{c|cccccccccccccccccccc|c}			
			\hline\hline
			Methods & C1 & C2 & C3 & C4 & C5 & C6 & C7 & C8 & C9 & C10 & C11 & C12 & C13 & C14 & C15 & C16 & C17 & C18 & C19 & C20 & mAP \\ \hline
			Baseline & 62.26 & 51.76 & 70.61 & 78.64 & 25.49 & 70.19 & 45.25 & 64.93 & 45.37 & 55.17 & 60.95 & 55.29 & 46.41 & 82.68 & 57.42 & 57.42 & 82.46 & 29.63 & 33.12 & 64.97 & 56.73 \\
			\makecell{Baseline + CSA-DRF} & 66.84 & \textbf{65.10} & 72.57 & 82.86 & 30.41 & 71.79 & 54.62 & 74.49 & 53.37 & 63.54 & 64.91 & 60.00 & 49.83 & 86.10 & 61.68 & 58.11 & 84.55 & \textbf{48.31} & 37.26 & 72.45 & 62.94 \\
			\textbf{LO-Det} & \textbf{72.63} & 65.04 & \textbf{76.72} & \textbf{84.66} & \textbf{33.46} & \textbf{73.71} & \textbf{56.83} & \textbf{75.86} & \textbf{57.51} & \textbf{66.29} & \textbf{68.01} & \textbf{60.91} & \textbf{51.50} & \textbf{88.63} & \textbf{68.04} & \textbf{64.31} & \textbf{86.26} & 47.57 & \textbf{42.44} & \textbf{76.70} & \textbf{65.85} \\
			\hline\hline
		\end{tabularx}}}
		{The explanation of each category}
		\resizebox{\textwidth}{!}{
			\renewcommand\arraystretch{1.5}
			\begin{tabular}{cccccccccccccccccccc}
				\hline\hline
				C1 & C2 & C3 & C4 & C5 & C6 & C7 & C8 & C9 & C10 & C11 & C12 & C13 & C14 &
				C15 & C16 & C17 & C18 & C19 & C20 \\ \hline
				Airplane & Airport & \makecell{Baseball\\ field} & \makecell{Basketball\\
					court} & Bridge & Chimney & Dam & \makecell{Expressway\\ service area} &
				\makecell{Expressway\\ toll station} & \makecell{Golf\\ course} & \makecell{Ground\\ track\\ field} & Harbor &
				Overpass & Ship & Stadium & \makecell{Storage\\ tank} & \makecell{Tennis\\ court} & \makecell{Train\\ station} &
				Vehicle & Windmill \\			
				\hline\hline
		\end{tabular}}
	\begin{tablenotes}
		\item Note: Bold font indicates the best results.
	\end{tablenotes}
	\end{threeparttable}
\end{table*}

\begin{table*}[htbp]
	\centering
	\setlength{\tabcolsep}{2.mm}{
	\caption{\label{table:4}
		{More detailed mAP (\%) results of ablation experiments on the DOTA dataset}}
	\begin{threeparttable}
		\centering
		\renewcommand\arraystretch{1.5}
		\resizebox{\linewidth}{!}{
			\begin{tabularx}{\textwidth}{c|ccccccccccccccc|c}
				\hline\hline
				Methods & PL & BD & BR & GTF & SV & LV & SH & TC & BC & ST & SBF & RA & HA & SP & HC & mAP \\ \hline
				Baseline1 & 88.50 & 69.45 & 29.26 & 41.49 & 68.96 & 69.26 & 81.05 & 90.63 & 64.92 & 78.88 & 38.81 & 53.88 & 58.71 & 57.65 & 37.13 & 61.91   \\
				Baseline2 & 88.68 & 65.99 & \textbf{32.80} & 45.02 & 68.93 & 69.64 & 81.43 & 90.57 & 66.40 & 79.06 & 44.20 & 56.25 & 59.19 & 65.88 & 28.87 & 62.86 \\
				Baseline3 & 88.71 & \textbf{69.14} & 31.48 & 51.16 & \textbf{70.64} & 70.46 & 83.52 & 90.68 & 69.44 & 77.90 & 41.49 & 58.55 & 58.02 & \textbf{66.62} & 32.65 & 64.03 \\
				\textbf{LO-Det} & \textbf{89.22} & 66.14 & 31.32 & \textbf{55.96} & 70.05 & \textbf{71.04} & \textbf{84.27} & \textbf{90.74} & \textbf{75.09} & \textbf{81.28} & \textbf{44.65} & \textbf{59.34} & \textbf{59.98} & 65.13 & \textbf{48.42} & 66.17\\
				\hline\hline
		\end{tabularx}}
		{The explanation of each category}
	\resizebox{\textwidth}{!}{
		\renewcommand\arraystretch{1.5}
		\begin{tabular}{ccccccccccccccc}
			\hline\hline
			PL & BD & BR & GTF & SV & LV & SH & TC & BC & ST & SBF & RA & HA & SP & HC \\ \hline
			Plane & \makecell{Baseball\\ diamond} & Bridge & \makecell{Ground field\\ track} & \makecell{Small\\ vehicle}
			& \makecell{Large\\ vehicle} & Ship & \makecell{Tennis\\ court} & \makecell{Basketball\\ court} & \makecell{Storage\\ tank}
			& \makecell{Soccer-ball\\ field} & Roundabout & Harbor & \makecell{Swimming\\ pool} & Helicopter
			\\			
			\hline\hline
	   \end{tabular}}
   \begin{tablenotes}
   	\item {Note: Bold font indicates the best results. Since the baseline method cannot detect OBBs, Baseline + gliding vertex \cite{xu2020gliding} is used as the new baseline. \\ Baseline1: Baseline + gliding vertex; Baseline2: Baseline + CSA-DRF + GV-Head; Baseline3: Baseline + CSA-DRF + Feature Balance + GV-Head.}
   \end{tablenotes}   

	\end{threeparttable}}		

\end{table*}

2) Evaluations of model parameters

It is mentioned in the analysis of the feature balance sub-module that the width of the model and the size of the input  image will also affect the performance of the model. There is usually a trade-off between accuracy and speed, which should be set carefully according to task requirements in different applications. The performance of the proposed LO-Det with various model widths and input image sizes on the DIOR \cite{liObjectDetectionOptical2020} dataset is evaluated as follows. 

The evaluation results of the model width are listed in Table~\ref{table:5}. LO-Det evaluated in the above experiments is represented by LO-Det × 1.0 with 1024, 512, and 256 feature maps in the three neck branches, respectively. Other different widths are adjusted based on this. For example, a CSA-DRF neck whose feature maps are half of the LO-Det × 1.0 is represented by LO-Det × 0.5. Table~\ref{table:5} evaluates the performances of these models. As the width of the CNN model decreases, the detection accuracy decreases, but the speed becomes faster and the model becomes smaller. This work provides the model design ideas and general model structure, and it can be customized according to the requirements to make a trade-off between accuracy and speed in applications. 

\begin{table*}[htbp]
	\centering
	\renewcommand\arraystretch{1.5}
	\caption{\label{table:5}
		{Evaluation of the proposed LO-Det with different model widths on the DIOR dataset}}
	\setlength{\tabcolsep}{1.5mm}{	
		\resizebox{\textwidth}{!}{
			\begin{tabular}{c|cccccccc}
				\hline\hline
				Models &  mAP (\%) & Speed 1  {(fps)} & Speed 2  {(fps)} & Speed 3  {(fps)} & Speed 4  {(fps)} & FLOPs & Parameters  {(MB)} & Model Size  {(MB)} \\ \hline
				LO-Det ×1.0 & 65.85 & 60.03 & 30.13 & 7.25 & 20.77 & 6.424G & 6.93 & 26.95 \\
				LO-Det ×0.75 & 51.99 & 67.52 & 37.48 & 11.45& 24.65 & 4.826G & 5.28 & 20.54 \\
				LO-Det ×0.5 & 43.87 & 69.60 & 40.00 & 13.67 & 29.11 & 3.618G & 3.97 & 15.53 \\
				LO-Det ×0.25 & 38.20 & 71.88 & 42.74 & 15.86 & 31.20 & 2.800G & 3.00 & 11.83 \\
				\hline\hline
		\end{tabular}}
		\begin{tablenotes}
			\item {Note: Speed 1 is the speed on RTX 3090 GPU, speed 2 is the speed on RTX 1660 GPU, speed 3 is the speed on NVIDIA Jetson TX2, Speed 4 is the \\ speed on NVIDIA Jetson AGX Xavier. LO-Det ×1.0 with 1024, 512, and 256 feature maps in three neck branches, respectively. Other different  widths \\ (×0.75, ×0.5, ×0.25) are adjusted based on this.}
		\end{tablenotes}}	
\end{table*}

\begin{table*}[htbp]
	\centering
	\renewcommand\arraystretch{1.5}
	\setlength{\tabcolsep}{1.0mm}{
		\caption{\label{table:6}
			{Evaluation of the proposed LO-Det with different input image sizes on the DIOR dataset}}
		\resizebox{\textwidth}{!}{
			\begin{tabular}{c|cccccccc}
				\hline\hline
				Models &  mAP (\%) & Speed 1  {(fps)} & Speed 2  {(fps)} & Speed 3  {(fps)} & Speed 4  {(fps)} & FLOPs & Parameters  {(MB)} & Model Size  {(MB)} \\ \hline
				LO-Det 608 & 65.85 & 60.03 & 30.13 & 7.25 & 20.77 & 6.424G & 6.93 & 26.95 \\
				LO-Det 512 & 64.06 & 62.87 & 33.15 & 9.08 & 22.75 & 4.556G & 6.93 & 26.95\\
				LO-Det 416 & 58.73 & 64.52 & 35.63 & 11.36 & 25.57 & 3.007G & 6.93 & 26.95 \\
				LO-Det 320 & 49.12 & 66.71 & 37.92 & 13.57 & 28.88 & 2.800G & 6.93 & 26.95 \\
				\hline\hline
		\end{tabular}}
		\begin{tablenotes}
			 \item {Note: Speed 1 is the speed on RTX 3090 GPU, speed 2 is the speed on RTX 1660 GPU, speed 3 is the speed on NVIDIA  Jetson TX2, Speed 4 is the \\ speed on NVIDIA Jetson AGX Xavier. LO-Det 608 indicates that the size of the input image (800×800 pixels) is down-sampled to 608×608 pixels (\\default value).}
		\end{tablenotes}
	}
\end{table*}

\begin{figure}[htbp]
	\centering
	\epsfig{width=0.49\textwidth,file=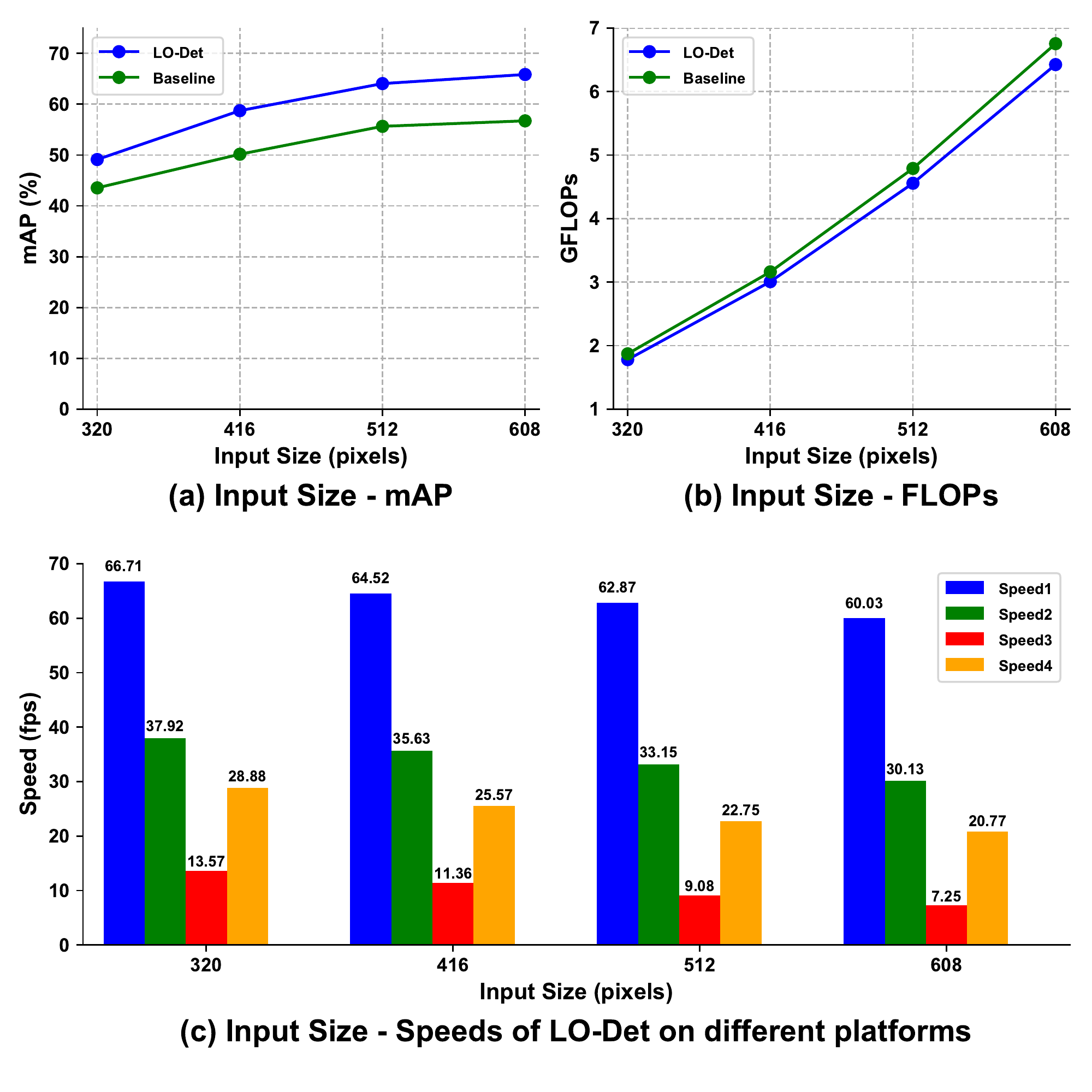}
	\caption{Performance analysis of the proposed LO-Det with different input sizes on the DIOR Dataset.}
	\label{fig:15}
\end{figure}

The evaluation results of the input image size are shown in Table~\ref{table:6} and Fig.~\ref{fig:15}. The effect of reducing the input image size on model performance is similar to that of reducing the width. As the input image size decreases, mAP decreases while the speed becomes faster. However, reducing the size of input images does not change the number of parameters and the model size, and the reduction in mAP is also less. 

Generally, in the case of fewer small objects or sufficient memory space, it is a better choice to reduce the input image size. In the case of limited speed and memory space, it is more suitable to reduce the model width in exchange for extremely fast speed and less memory consumption. In addition to MobileNetv2 \cite{sandler2018mobilenetv2}, the performance of LO-Det using other lightweights, such as ShuffleNetv2 \cite{ma2018shufflenet} and GhostNet \cite{han2020ghostnet}, is also evaluated in Table~\ref{table:7}. From the experimental results in Table~\ref{table:7}, using MobileNetv2 \cite{sandler2018mobilenetv2} as the backbone makes the proposed LO-Det perform better than using ShuffleNetv2 \cite{ma2018shufflenet} or GhostNet \cite{han2020ghostnet} as the backbone on more evaluation metrics, so MobileNetv2 \cite{sandler2018mobilenetv2} is chosen as the backbone of LO-Det.

\begin{table*}[htbp]
	\centering
	\renewcommand\arraystretch{1.5}
	\setlength{\tabcolsep}{1.0mm}{
		\caption{\label{table:7}
			{Evaluation of the proposed LO-Det with different backbone networks on the DIOR dataset}}
		\resizebox{\textwidth}{!}{
			\begin{tabular}{c|cccccccc}
				\hline\hline
				Models &  mAP & Speed 1  {(fps)} & Speed 2 {(fps)} & Speed 3  {(fps)} & Speed 4  {(fps)} & FLOPs & Parameters  {(MB)} & Model Size  {(MB)} \\ \hline
				LO-Det (MobileNetv2 ×1.0 \cite{sandler2018mobilenetv2}) & 65.85 & 60.03 & 30.13 & 7.25 & 20.77 & 6.424G & 6.93 & 26.95 \\
				LO-Det (ShuffleNetv2 ×1.5 \cite{ma2018shufflenet}) & 65.99 & 59.32 & 28.81 & 6.97 & 20.15 & 6.922G & 9.39 & 36.25\\
				LO-Det (GhostNetv2 ×1.0 \cite{han2020ghostnet}) & 66.01 & 40.71 & 16.61 & 4.39 & 12.88 & 5.0822G & 7.14 & 27.76 \\
				\hline\hline
		\end{tabular}}
		\begin{tablenotes}
			\item {Note: MobileNetv2 ×1.0 with 1280, 96, and 32 feature maps input to the three neck branches, respectively, as shown  in Fig.~\ref{fig:2}. ShuffleNetv2 ×1.5 with \\ 1024, 352, 176 feature maps input to the three neck branches, respectively. And  GhostNetv2 ×1.0 with 960, 112, and 40 feature maps input to the three \\ neck  branches, respectively. The details of these  lightweight backbones are explained in \cite{sandler2018mobilenetv2}, \cite{ma2018shufflenet}, and \cite{han2020ghostnet}.}
		\end{tablenotes}
	}
\end{table*}

\subsection{Comparative Experiments and Discussions}

Due to lacks of benchmarking methods, the performance of the proposed LO-Det is compared with the much larger models with the state-of-the-art mAP. Even if this is unfair to the proposed LO-Det, it is also helpful to discover the shortcomings of the existing lightweight object detectors and future improvements. The comparative experiments are conducted on the DIOR dataset and DOTA dataset for comprehensively evaluation.

1) Comparative experiments on the DIOR dataset

\begin{figure}[htbp]
	\centering
	\epsfig{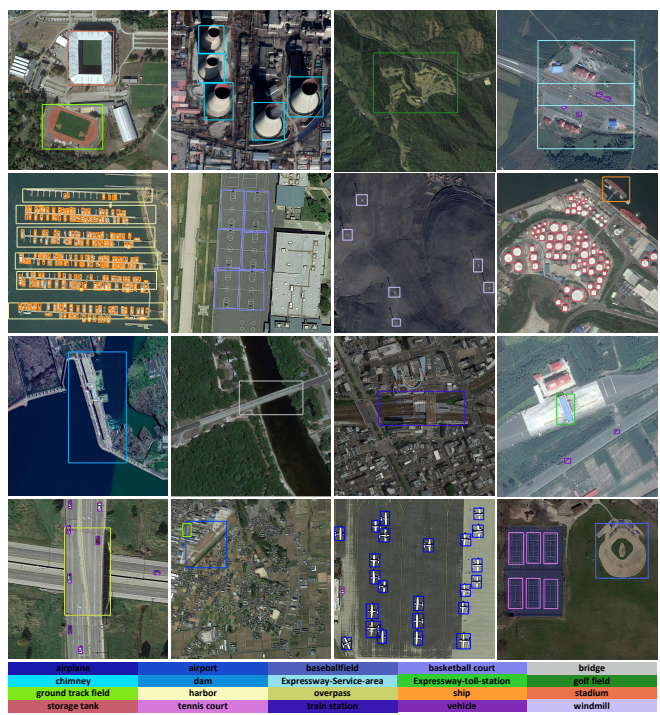}
	\caption{Visualization Results of LO-Det on the DIOR Dataset.}
	\label{fig:16}
\end{figure}

The results of comparative experiments on the DIOR dataset are listed in Table~\ref{table:8}, and the visual results are shown in Fig.~\ref{fig:16} intuitively. The proposed LO-Det has a mAP of 65.9, which is higher than that of Faster R-CNN (R50) \cite{renFasterRCNNRealTime2017a}, SSD \cite{liuSSDSingleShot2016a}, RetinaNet (R50) \cite{linFocalLossDense2017}, CornerNet \cite{lawCornerNetDetectingObjects2018} and many mainstream methods and is only slightly lower than that of CSFF \cite{chengCrossScaleFeatureFusion2020c}. Furthermore, the proposed LO-Det has a very fast detection speed and very small memory consumption. It can reach a detection speed of 60.03 fps on an RTX 3090 GPU when the input image size is 608×608 pixels, and only takes up 26.95 MB of memory consumption. This detection speed is 4-5 times faster than that of Faster R-CNN (R50) \cite{renFasterRCNNRealTime2017a}, RetinaNet (R50) \cite{linFocalLossDense2017}, CornerNet \cite{lawCornerNetDetectingObjects2018}, Mask R-CNN \cite{heMaskRCNN2017b}, and PANet \cite{wangPANetFewShotImage2019}. It is also faster than that of SSD \cite{liuSSDSingleShot2016a}, and the storage space required for this model is only one-fifth or less of the storage space of other methods. In order to pursue higher detection accuracy, LO-Det 800 ×1.5 is designed, in which the CSA-DRF neck becomes 1.5 times wider and the size of input images is 800×800 pixels. This model reaches the state-of-the-art mAP of 68.5 at 58.27 fps and only need 43.41 MB for storage.

\begin{table*}[htbp]
	\centering
	\renewcommand\arraystretch{2}
	\caption{\label{table:8}
		{Comparative experiments of mAP (\%) and speed (fps) on the DIOR dataset}}
	\setlength{\tabcolsep}{0.72mm}
	\begin{threeparttable}
		\centering
		\resizebox{\textwidth}{!}{
			\begin{tabularx}{\textwidth}{c|c|cccccccccccccccccccc|c|c|c}
				\hline\hline
				Methods & Year & C1 & C2 & C3 & C4 & C5 & C6 & C7 & C8 & C9 & C10 & C11 & C12 & C13 & C14 & C15 & C16 & C17 & C18 & C19 & C20 &\makecell{ mAP \\ (\%)}& \makecell{Speed 1 \\ (fps)} & \makecell{Model \\ Size \\ (MB)} \\ \hline
				\makecell{Faster R-CNN \\ (R50) \cite{renFasterRCNNRealTime2017a}} & 2016 & 54.1 & 71.4 & 63.3 & 81.0 & 42.6 & 72.5 & 57.5 & 68.7 & 62.1 & 73.1 & 76.5 & 42.8 & 56.0 & 71.8 & 57.0 & 53.5 & 81.2 & 53.0 & 43.1 & 80.9 & 63.1 & 19.91 & 161.20 \\ \hline
				SSD \cite{liuSSDSingleShot2016a} & 2016 & 59.5 & 72.7 & 72.4 & 75.7 & 29.7 & 65.8 & 56.6 & 63.5 & 53.1 & 65.3 & 68.6 & 49.4 & 48.1 & 59.2 & 61.0 & 46.6 & 76.3 & 55.1 & 27.4 & 65.7 & 58.6 & 54.18 & 169.68 \\ \hline
				\makecell{Mask R-CNN \\ (R50) \cite{heMaskRCNN2017b}} & 2017 & 53.8 & 72.3 & 63.2 & 81.0 & 38.7 & 72.6 & 55.9 & 71.6 & 67.0 & 73.0  & 75.8 & 44.2 & 56.5 & 71.9 & 58.6 & 53.6 & 81.1 & 54.0 & 43.1 & 81.1 & 63.5 & 16.33 & 130.86 \\ \hline
				\makecell{RetinaNet \\ (R50) \cite{linFocalLossDense2017}} & 2017 & 53.7 & 77.3 & 69.0 & 81.3 & 44.1 & 72.3 & 62.5 & 76.2 & 66.0 & 77.7 & 74.2 & 50.7 & 59.6 & 71.2 & 69.3 & 44.8 & 81.3 & 54.2 & 44.4 & 83.4 & 65.7 & 18.26 & 145.03 \\ \hline
				CornerNet \cite{lawCornerNetDetectingObjects2018} & 2018 & 58.8  & \textbf{84.2}  & 72.0  &  80.8 & \textbf{46.4}  & 75.3  & \textbf{64.3}  & 81.6  & \textbf{76.3}  & \textbf{79.5}  & 79.5  &  26.1 & \textbf{60.6}  & 37.6  &  70.7 & 45.2  & 84.0  & 57.1  & 43.0  & 75.9  & 64.9  &  10.78 & 768.17  \\ \hline
				\makecell{PANet \\ (R50) \cite{wangPANetFewShotImage2019}} & 2018 & 61.9  & 70.4  & 71.0  & 80.4  & 38.9  & 72.5  & 56.6  & 68.4  & 60.0  &  69.0 &  74.6 & 41.6  &  55.8 &  71.7 & 72.9  & 62.3  & 81.2 & 54.6	 & 48.2  & 86.7  & 63.8  & 10.88  & 172.97  \\	\hline	 			
				CSFF \cite{chengCrossScaleFeatureFusion2020c} & 2020 & 57.2  &  79.6 & 70.1  &  87.4 &  46.1 & \textbf{76.6}  &  62.7 & \textbf{82.6}  & 73.2  & 78.2  &  \textbf{81.6} & 50.7  & 59.5  & 73.3  & 63.4  & 58.5  & 85.9  & \textbf{61.9}  & 42.9  &  \textbf{86.9} & 68.0  &  15.21 &  168.71 \\ \hline										
				\textbf{LO-Det 608} & 2020 & 72.6  & 65.0  & 76.7  & 84.7  & 33.5  & 73.7  &  56.8 & 75.9  & 57.5  &  66.3 & 68.0  & \textbf{60.9}  & 51.5  &  88.6 & 68.0  & 64.3  &  86.3 & 47.6  & 42.4  & 76.7  & 65.9  & \textbf{60.03}  & \textbf{26.95} \\ \hline				
				\textbf{\makecell{LO-Det 800 \\ ×1.5}} & 2020 & \textbf{77.5}  & 60.9  & \textbf{80.3}  &  \textbf{88.0} &  36.7 & 75.0  & 54.0 &  74.7 & 63.2  &  63.4 & 73.0  & 60.6  & 53.8  & \textbf{90.4}  & \textbf{74.6} & \textbf{77.3}  &  \textbf{88.6} & 43.6  & \textbf{51.1}  & 84.0 & \textbf{68.5} & 58.27  & 43.41 \\
				\hline\hline
		\end{tabularx}}
		{Explanation of each category}
		\resizebox{\textwidth}{!}{
			\renewcommand\arraystretch{1.5}
			\setlength{\tabcolsep}{0.5mm}
		\begin{tabular}{cccccccccccccccccccc}
				\hline\hline
				C1 & C2 & C3 & C4 & C5 & C6 & C7 & C8 & C9 & C10 & C11 & C12 & C13 & C14 &
				C15 & C16 & C17 & C18 & C19 & C20 \\ \hline
				Airplane & Airport & \makecell{Baseball\\ field} & \makecell{Basketball\\
					court} & Bridge & Chimney & Dam & \makecell{Expressway\\ service area} &
				\makecell{Expressway\\ toll station} & \makecell{Golf\\ course} & \makecell{Ground\\ track\\ field} & Harbor &
				Overpass & Ship & Stadium & \makecell{Storage\\ tank} & \makecell{Tennis\\ court} & \makecell{Train\\ station} &
				Vehicle & Windmill \\			
				\hline\hline
		\end{tabular}}
		\begin{tablenotes}
			\item{
					Note: Bold font indicates the best results. Since the benchmarking methods and their codes are not designed for embedded devices, only the speed on \\ an RTX 3090 GPU (Speed 1) is evaluated. In addition, some methods  are re-implemented by ourselves, the results may be slightly different from  those\\ reported in the original papers. The size of images used by the compared methods is also 800×800 pixels.
				}
		\end{tablenotes}
	\end{threeparttable}
\end{table*}

In addition, the proposed LO-Det has low accuracy for detecting large objects, such as airport, golf course, and train station. This is due to that the lightweight network's ability to learn deep semantic features is not as good as the large CNN model, resulting in low detection accuracy of large objects with rich texture and detailed features.

2) Comparative experiments on the DOTA dataset

\begin{figure*}[htbp]
	\centering
	\epsfig{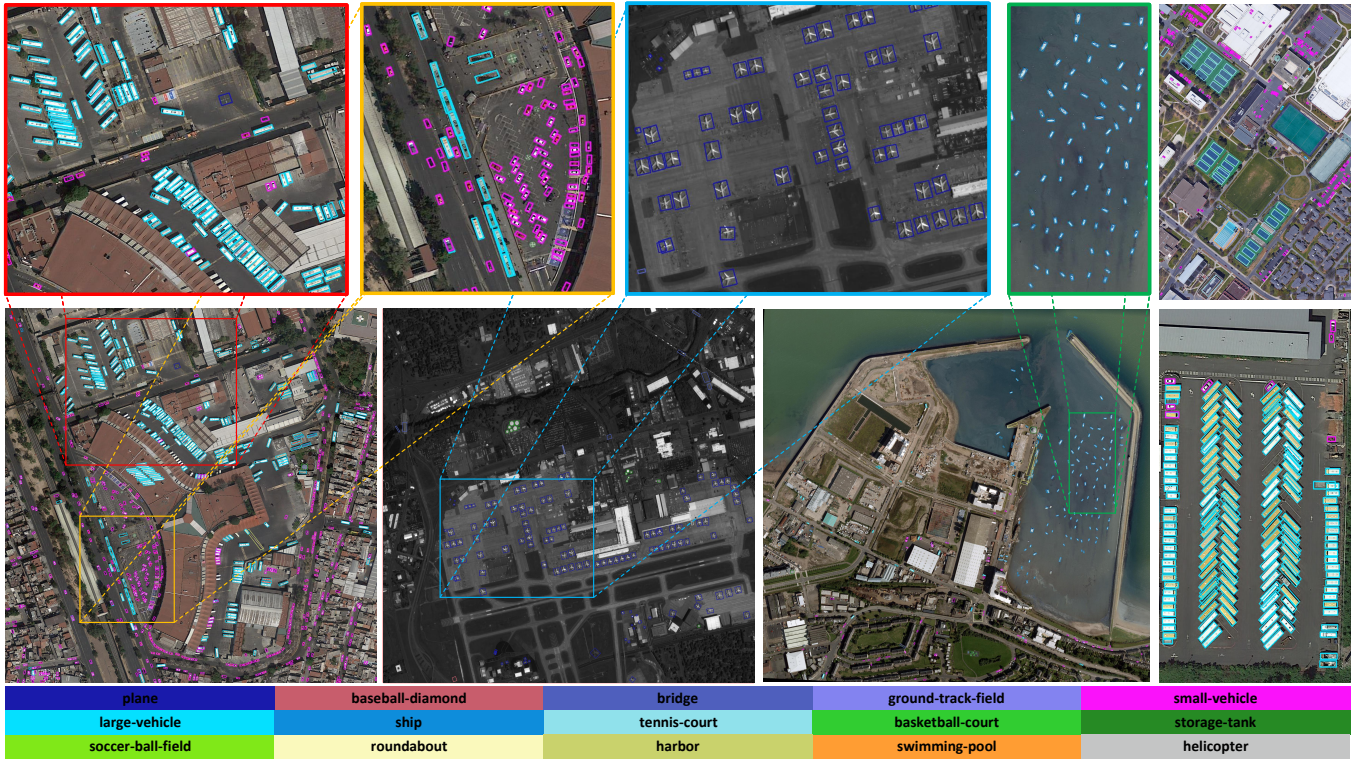}
	\caption{Visualization Results of LO-Det on the DOTA Dataset.}
	\label{fig:17}
\end{figure*}

The results of comparative experiments are listed in Table~\ref{table:9}, and the visual results are shown in Fig.~\ref{fig:17} intuitively. The proposed LO-Det has a mAP of 66.17, which is higher than that of YOLOv2 \cite{redmon2017yolo9000}, SSD \cite{liuSSDSingleShot2016a}, FR-O \cite{xiaDOTALargeScaleDataset2018}, and R${^2}$CNN \cite{jiang2017r2cnn}. Although the detection accuracy of the proposed LO-Det is lower than that of extremely large CNN models, such as CAD-Net \cite{zhangCADNetContextAwareDetection2019}, SCRDet \cite{yangSCRDetMoreRobust2019}, Gliding Vertex \cite{xu2020gliding}, etc., its detection speed is 5 times faster than that of these large models, and its model size is only one-tenth or less of these large models. Moreover, the proposed LO-Det can run on embedded devices, and its speed on embedded devices is even faster than that of those large models running on servers with expensive GPUs, such as RTX 3090. In order to pursue the ultimate high accuracy, LO-Det 800 (Vanilla Conv) is also designed according to the proposed idea, in which Darknet53 (a large and complex CNN model) is used as the backbone network, to compare with the Gliding Vertex (Faster R-CNN R101) \cite{xu2020gliding}, which is very large. The proposed LO-Det 800 (Vanilla Conv) reaches a mAP of 75.24 at 27.44 fps, the performance of which is better than that of Gliding Vertex (Faster R-CNN R101) \cite{xu2020gliding}. 

\begin{table*}[htbp]
	\centering
	\renewcommand\arraystretch{2}
	\caption{\label{table:9}
		{Comparative experiments of mAP (\%) and speed (fps) on the DOTA dataset}}
	\setlength{\tabcolsep}{0.85mm}
	\begin{threeparttable}
		\centering
		\resizebox{\textwidth}{!}{
			\begin{tabularx}{\textwidth}{c|c|ccccccccccccccc|c|c|c}
				\hline\hline
				Methods & Year & PL & BD & BR & GTF & SV & LV & SH & TC & BC & ST & SBF & RA & HA & SP & HC & \makecell{mAP \\ (\%)} & \makecell{Speed 1 \\ (fps)} & \makecell{Model \\ Size \\ (MB)} \\ \hline
				YOLOv2 \cite{redmon2017yolo9000} & 2016 & 39.57 & 20.29 & 36.58 & 23.42 & 8.85 & 2.09 & 4.82 & 44.34 & 38.35 & 34.65 & 16.02 & 37.62 & 47.23 & 25.50 & 7.45 & 21.39 & 47.52 & 192.46 \\ \hline
				SSD \cite{liuSSDSingleShot2016a} & 2016 & 39.83 & 9.09 & 0.64 & 13.18 & 0.26 & 0.39 & 1.11 & 16.24 & 27.57 & 9.23 & 27.16 & 9.09 & 3.03 & 1.05 & 1.01 & 10.59 & 53.20 & 170.00 \\			\hline							
				FR-O \cite{xiaDOTALargeScaleDataset2018} & 2018 & 79.42 & 77.13 & 17.70 & 64.05 & 35.30 & 38.02 & 37.13 & 89.41 & 69.64 & 59.28 & 50.30 & 52.91 & 47.89 & 47.40 & 46.30 & 54.13 & - & - \\ \hline
				R$^{2}$CNN \cite{jiang2017r2cnn} & 2018 & 80.94 & 65.67 & 35.34 & 67.44 & 59.92 & 50.91 & 55.81 & 90.67 & 72.39  & 66.92 & 55.06 & 52.23 & 55.14 & 53.35 & 48.22 & 60.67 & 13.02 & 170.81 \\ \hline
				ROI Trans. \cite{xiaDOTALargeScaleDataset2018} & 2019 & 88.53 & 77.91 & 37.63 & 74.08 & 66.53 & 62.97 & 66.57 & 90.50 & 79.46 & 76.75 & 59.04 & 56.73 & 62.54 & 61.29 & 55.56 & 67.74 & 7.10 & 273.20 \\ \hline
				CAD-Net \cite{zhangCADNetContextAwareDetection2019} & 2019 & 87.80 & 82.40 & 49.40 & 73.50  & 71.10 & 63.50 & 76.60 & \textbf{90.90} & 79.20 & 73.30 & 48.40 & 60.90 & 62.00 & 67.00 & 62.20 & 72.72 & - & - \\ \hline
				SCRDet \cite{yangSCRDetMoreRobust2019} & 2019 & \textbf{89.98} & 80.65 & 52.09 & 68.36  & 68.36 & 60.32 & 72.41 & 90.85 & \textbf{87.94} & \textbf{86.86} & \textbf{65.02} & 66.68 & 66.25 & 68.24 & \textbf{65.21} & 72.61 & 7.40 & 338.90 \\ \hline	
				\makecell{Gliding Vertex \cite{xu2020gliding}} & 2019 & 89.64 & \textbf{85.00} & \textbf{52.26} & \textbf{77.34} & 73.01 & 73.14 & 86.82 & 90.74 & 79.02 & 86.81 & 59.55 & \textbf{70.91} & 72.94 & 70.86 & 57.32 & 75.02 & 13.10 & 460.90 \\ \hline
				FFA3 \cite{fuRotationawareMultiscaleConvolutional2020} & 2020 & 88.80 & 74.40 & 48.90 & 57.90 & 63.60 & 75.90 & 79.60 & 90.80 & 80.30 & 82.90 & 54.30 & 60.00 & 66.90 & 66.80 & 42.50 & 68.90 & - & - \\ \hline			
				\textbf{LO-Det 608} & 2020 & 89.22 & 66.14 & 31.32 & 55.96 & 70.05 & 71.04 & 84.27 & 90.74 & 75.09 & 81.28 & 44.65 & 59.34 & 59.98 & 65.13 & 48.42 & 66.17 & \textbf{60.01} & \textbf{26.95} \\ \hline
				\textbf{\makecell{LO-Det 800 \\ (Vanilla Conv)}} & 2020 &  89.91 & 84.92  & 47.04  & 69.53  &  \textbf{75.80} &  \textbf{76.37} &  \textbf{88.48} & 90.87  & 86.25 & 86.58  & 62.39 & 68.15  & \textbf{73.18} & \textbf{71.59}  &  57.61 &  \textbf{75.24} & 27.44  & 397.91 \\ 
				\hline\hline
		\end{tabularx}}
		{The explanation of each category}
		\resizebox{\textwidth}{!}{
			\setlength{\tabcolsep}{1.17mm}
			\renewcommand\arraystretch{1.5}
			\begin{tabular*}{\textwidth}{ccccccccccccccc}
				\hline\hline
				PL & BD & BR & GTF & SV & LV & SH & TC & BC & ST & SBF & RA & HA & SP & HC \\ \hline
				Plane & \makecell{Baseball\\ diamond} & Bridge & \makecell{Ground field\\ track} & \makecell{Small\\ vehicle}
				& \makecell{Large\\ vehicle} & Ship & \makecell{Tennis\\ court} & \makecell{Basketball\\ court} & \makecell{Storage\\ tank}
				& \makecell{Soccer-ball\\ field} & Roundabout & Harbor & \makecell{Swimming\\ pool} & Helicopter
				\\
				\hline\hline
		\end{tabular*}}
		\begin{tablenotes}
			\item {Bold font indicates the best results. Since the benchmarking methods and their codes are not designed for embedded devices, only the speed on the single RTX 3090 GPU (Speed 1) is evaluated. Additionally, some methods are re-implemented by ourselves, the results may be slightly different from those reported in the original papers. And some methods’ codes are not open-source, we cannot test their detection speed, which is indicated by “-”.}
		\end{tablenotes}
	\end{threeparttable}
\end{table*}

In addition, comparing the proposed LO-Det with the large models, the performance gap between them is mainly in the detection of ground track field, soccer-ball field, harbor and other large objects, which is consistent with the results on the DIOR dataset.

\section{Conclusions}
In this paper, a novel lightweight detector LO-Det is proposed for oriented object detection in remote sensing images. It is the first lightweight detector designed for RSOD tasks. In LO-Det, the CSA-DRF neck component is designed to optimize detection efficiency and accuracy by the channel separation-aggregation structure and dynamic receptive field mechanism. The DSC-Head component is developed to detect OBBs and constrain the shapes of OBBs more accurately through the diagonal support constraint and the M-Sigmoid function.

The experiments demonstrate the following: 1) Each component proposed in LO-Det is valid, in which the CSA-DRF reduces model complexity and improves detection accuracy, the DSC-Head makes the shapes of OBBs more accurate and alleviates the boundary value problem. 2) The detection accuracy of LO-Det is not inferior to the mainstream methods while LO-Det is much faster than them. 3) The proposed LO-Det only needs a single GPU to train. It can not only perform inferences on expensive GPUs like RTX 3090, but also run fast on embedded devices. 4) The proposed LO-Det has great potential for development, which can adapt to different RSOD tasks in different scenarios and on different platforms by adjusting the model structure and parameters based on the explored rules. 

Despite its demonstrated benefits, the detection accuracy of LO-Det for large objects needs to be improved, and the sparse detection will be studied for higher efficiency in the future. In addition, more comprehensive optimization of the backbone, the neck component, and many other operations may obtain better detection performance.

The source code and models of this work will be available at https://github.com/Shank2358. Besides, a model zoo containing LO-Det using other lightweight backbones will be also available.


\small
\bibliographystyle{IEEEtran}
\bibliography{LODet.bib}

\begin{thebibliography}{10}
\providecommand{\url}[1]{#1}
\csname url@rmstyle\endcsname
\providecommand{\newblock}{\relax}
\providecommand{\bibinfo}[2]{#2}
\providecommand\BIBentrySTDinterwordspacing{\spaceskip=0pt\relax}
\providecommand\BIBentryALTinterwordstretchfactor{4}
\providecommand\BIBentryALTinterwordspacing{\spaceskip=\fontdimen2\font plus
\BIBentryALTinterwordstretchfactor\fontdimen3\font minus
  \fontdimen4\font\relax}
\providecommand\BIBforeignlanguage[2]{{%
\expandafter\ifx\csname l@#1\endcsname\relax
\typeout{** WARNING: IEEEtran.bst: No hyphenation pattern has been}%
\typeout{** loaded for the language `#1'. Using the pattern for}%
\typeout{** the default language instead.}%
\else
\language=\csname l@#1\endcsname
\fi
#2}}

\bibitem{renFasterRCNNRealTime2017a}
S.~Ren, K.~He, R.~Girshick, and J.~Sun, ``Faster {{R}}-{{CNN}}: {{Towards
  Real}}-{{Time Object Detection}} with {{Region Proposal Networks}},''
  \emph{IEEE Transactions on Pattern Analysis and Machine Intelligence},
  vol.~39, no.~6, pp. 1137--1149, 2017.

\bibitem{liuSSDSingleShot2016a}
W.~Liu, D.~Anguelov, D.~Erhan, C.~Szegedy, S.~E. Reed, C.-Y. Fu, and A.~C.
  Berg, ``{{SSD}}: {{Single Shot MultiBox Detector}},'' in \emph{Proceedings of
  {{European Conference}} on {{Computer Vision}}}, {Amsterdam, The
  Netherlands}, Oct. 2016, pp. 21--37.

\bibitem{redmonYOLOv3IncrementalImprovement2018}
J.~Redmon and A.~Farhadi, ``{{YOLOv3}}: {{An Incremental Improvement}},''
  \emph{arXiv preprint arXiv:1804.02767}, 2018.

\bibitem{chengLearningRotationInvariantConvolutional2016}
G.~Cheng, P.~Zhou, and J.~Han, ``Learning {{Rotation}}-{{Invariant
  Convolutional Neural Networks}} for {{Object Detection}} in {{VHR Optical
  Remote Sensing Images}},'' \emph{IEEE Transactions on Geoscience and Remote
  Sensing}, vol.~54, no.~12, pp. 7405--7415, 2016.

\bibitem{yangSCRDetMoreRobust2019}
X.~Yang, J.~Yang, J.~Yan, Y.~Zhang, T.~Zhang, Z.~Guo, X.~Sun, and K.~Fu,
  ``{{SCRDet}}: {{Towards More Robust Detection}} for {{Small}}, {{Cluttered}}
  and {{Rotated Objects}},'' in \emph{Proceedings of {{IEEE International
  Conference}} on {{Computer Vision}}}, {Seoul, South Korea}, Oct. 2019, pp.
  8232--8241.

\bibitem{fu2020point}
K.~Fu, Z.~Chang, Y.~Zhang, and X.~Sun, ``Point-based {{Estimator}} for
  {{Arbitrary}}-{{Oriented Object Detection}} in {{Aerial Images}},''
  \emph{IEEE Transactions on Geoscience and Remote Sensing}, pp. 1--18, 2020.

\bibitem{huang2020dc}
Z.~Huang, J.~Wang, X.~Fu, T.~Yu, Y.~Guo, and R.~Wang,
  ``{{DC}}-{{SPP}}-{{YOLO}}: {{Dense}} connection and spatial pyramid pooling
  based {{YOLO}} for object detection,'' \emph{Information Sciences}, vol. 522,
  pp. 241--258, 2020.

\bibitem{linFeaturePyramidNetworks2017a}
T.-Y. Lin, P.~Dollar, R.~Girshick, K.~He, B.~Hariharan, and S.~Belongie,
  ``Feature {{Pyramid Networks}} for {{Object Detection}},'' in
  \emph{Proceedings of {{IEEE Conference}} on {{Computer Vision}} and {{Pattern
  Recognition}}}, {Honolulu, Hawaii, USA}, July 2017, pp. 936--944.

\bibitem{szegedy2016rethinking}
C.~Szegedy, V.~Vanhoucke, S.~Ioffe, J.~Shlens, and Z.~Wojna, ``Rethinking the
  {{Inception Srchitecture}} for {{Computer Vision}},'' in \emph{Proceedings of
  {{IEEE International Conference}} on {{Computer Vision}}}, {Las Vegas, NV,
  USA}, June 2016, pp. 2818--2826.

\bibitem{wangFMSSDFeatureMergedSingleShot2020}
P.~Wang, X.~Sun, W.~Diao, and K.~Fu, ``{{FMSSD}}: {{Feature}}-{{Merged
  Single}}-{{Shot Detection}} for {{Multiscale Objects}} in {{Large}}-{{Scale
  Remote Sensing Imagery}},'' \emph{IEEE Transactions on Geoscience and Remote
  Sensing}, vol.~58, no.~5, pp. 3377--3390, 2020.

\bibitem{fuRotationawareMultiscaleConvolutional2020}
K.~Fu, Z.~Chang, Y.~Zhang, G.~Xu, K.~Zhang, and X.~Sun, ``Rotation-aware and
  {{Multi}}-{{Scale Convolutional Neural Network}} for {{Object Detection}} in
  {{Remote Sensing Images}},'' \emph{ISPRS Journal of Photogrammetry and Remote
  Sensing}, vol. 161, pp. 294--308, 2020.

\bibitem{huSqueezeandExcitationNetworks2018}
J.~Hu, L.~Shen, and G.~Sun, ``Squeeze-and-{{Excitation Networks}},'' in
  \emph{Proceedings of {{IEEE Conference}} on {{Computer Vision}} and {{Pattern
  Recognition}}}, {Salt Lake City, Utah, USA}, June 2018, pp. 7132--7141.

\bibitem{wooCBAMConvolutionalBlock2018}
S.~Woo, J.~Park, J.-Y. Lee, and I.~S. Kweon, ``{{CBAM}}: {{Convolutional Block
  Attention Module}},'' in \emph{Proceedings of the {{European Conference}} on
  {{Computer Vision}}}, {Munich, Germany}, Sept. 2018, pp. 3--19.

\bibitem{zhangCADNetContextAwareDetection2019}
G.~Zhang, S.~Lu, and W.~Zhang, ``{{CAD}}-{{Net}}: {{A Context}}-{{Aware
  Detection Network}} for {{Objects}} in {{Remote Sensing Imagery}},''
  \emph{IEEE Transactions on Geoscience and Remote Sensing}, vol.~57, no.~12,
  pp. 10\,015--10\,024, 2019.

\bibitem{9364888}
Z.~Huang, W.~Li, X.~G. Xia, X.~Wu, Z.~Cai, and R.~Tao, ``A novel nonlocal-aware
  pyramid and multiscale multitask refinement detector for object detection in
  remote sensing images,'' \emph{IEEE Transactions on Geoscience and Remote
  Sensing}, pp. 1--20, 2021.

\bibitem{yangSCRDetDetectingSmall2020b}
X.~Yang, J.~Yan, X.~Yang, J.~Tang, W.~Liao, and T.~He, ``{{SCRDet}}++:
  {{Detecting Small}}, {{Cluttered}} and {{Rotated Objects}} via
  {{Instance}}-{{Level Feature Denoising}} and {{Rotation Loss Smoothing}},''
  \emph{arXiv preprint arXiv:2004.13316}, 2020.

\bibitem{lin2019ienet}
Y.~Lin, P.~Feng, and J.~Guan, ``{{IENet}}: {{Interacting Embranchment One Stage
  Anchor Free Detector}} for {{Orientation Aerial Object Detection}}.''
  \emph{arXiv preprint arXiv:1912.00969}, 2019.

\bibitem{howard2017mobilenets}
A.~G. Howard, M.~Zhu, B.~Chen, D.~Kalenichenko, W.~Wang, T.~Weyand,
  M.~Andreetto, and H.~Adam, ``{{MobileNets}}: {{Efficient Convolutional Neural
  Networks}} for {{Mobile Vision Applications}},'' \emph{arXiv preprint
  arXiv:1704.04861}, 2017.

\bibitem{sandler2018mobilenetv2}
M.~Sandler, A.~Howard, M.~Zhu, A.~Zhmoginov, and L.-C. Chen, ``{{MobileNetV2}}:
  {{Inverted Residuals}} and {{Linear Bottlenecks}},'' in \emph{Proceedings of
  {{IEEE Conference}} on {{Computer Vision}} and {{Pattern Recognition}}},
  {Salt Lake City, Utah, USA}, 2018, pp. 4510--4520.

\bibitem{howard2019searching}
A.~Howard, R.~Pang, H.~Adam, Q.~Le, M.~Sandler, B.~Chen, W.~Wang, L.-C. Chen,
  M.~Tan, G.~Chu, V.~Vasudevan, and Y.~Zhu, ``Searching for {{MobileNetV3}},''
  in \emph{Proceedings of {{IEEE International Conference}} on {{Computer
  Vision}}}, {Seoul, South Korea}, Oct. 2019, pp. 1314--1324.

\bibitem{zhang2018shufflenet}
X.~Zhang, X.~Zhou, M.~Lin, and J.~Sun, ``{{ShuffleNet}}: {{An Extremely
  Efficient Convolutional Neural Network}} for {{Mobile Devices}},'' in
  \emph{Proceedings of {{IEEE Computer Vision}} and {{Pattern Recognition}}},
  {Salt Lake City, Utah, USA}, June 2018, pp. 6848--6856.

\bibitem{ma2018shufflenet}
N.~Ma, X.~Zhang, H.-T. Zheng, and J.~Sun, ``{{ShuffleNet V2}}: {{Practical
  Guidelines}} for {{Efficient CNN Architecture Design}},'' in
  \emph{Proceedings of the {{European Conference}} on {{Computer Vision}}},
  {Munich, Germany}, Sept. 2018, pp. 122--138.

\bibitem{han2020ghostnet}
K.~Han, Y.~Wang, Q.~Tian, J.~Guo, C.~Xu, and C.~Xu, ``{{GhostNet}}: {{More
  Features}} from {{Cheap Operations}},'' in \emph{Proceedings of {{IEEE
  Conference}} on {{Computer Vision}} and {{Pattern Recognition}}}, {Online},
  2020, pp. 1580--1589.

\bibitem{wang2018pelee}
R.~J. Wang, X.~Li, and C.~X. Ling, ``Pelee: {{A Real}}-{{Time Object Detection
  System}} on {{Mobile Devices}},'' in \emph{Proceedings of {{International
  Conference}} on {{Neural Information Processing Systems}}}, vol.~31,
  {Montr\'eal, Canada}, Dec. 2018, pp. 1967--1976.

\bibitem{yang2019legonet}
Z.~Yang, Y.~Wang, C.~Liu, H.~Chen, C.~Xu, B.~Shi, C.~Xu, and C.~Xu,
  ``{{LegoNet}}: {{Efficient Convolutional Neural Networks}} with {{Lego
  Filters}},'' in \emph{International Conference on Machine Learning}, {Long
  Beach, CA, USA}, June 2019, pp. 7005--7014.

\bibitem{chollet2017xception}
F.~Chollet, ``Xception: {{Deep Learning}} with {{Depthwise Separable
  Convolutions}},'' in \emph{Proceedings of {{IEEE Computer Vision}} and
  {{Pattern Recognition}}}, {Honolulu, Hawaii, USA}, July 2017, pp. 1800--1807.

\bibitem{iandola2016squeezenet}
F.~N. Iandola, S.~Han, M.~W. Moskewicz, K.~Ashraf, W.~J. Dally, and K.~Keutzer,
  ``{{SqueezeNet}}: {{AlexNet}}-{{Level Accuracy}} with 50x {{Fewer
  Parameters}} and {$<$}0.{{5MB Model Size}},'' \emph{arXiv preprint
  arXiv:1602.07360}, 2016.

\bibitem{liLightHeadRCNNDefense2017a}
Z.~Li, C.~Peng, G.~Yu, X.~Zhang, Y.~Deng, and J.~Sun, ``Light-{{Head
  R}}-{{CNN}}: {{In Defense}} of {{Two}}-{{Stage Object Detector}}.''
  \emph{arXiv preprint arXiv:1711.07264}, 2017.

\bibitem{qin2019thundernet}
Z.~Qin, Z.~Li, Z.~Zhang, Y.~Bao, G.~Yu, Y.~Peng, and J.~Sun, ``{{ThunderNet}}:
  {{Towards Real}}-{{Time Generic Object Detection}} on {{Mobile Devices}},''
  in \emph{Proceedings of {{IEEE International Conference}} on {{Computer
  Vision}}}, {Seoul, South Korea}, Oct. 2019, pp. 6718--6727.

\bibitem{tang2020lightdet}
Q.~Tang, J.~Li, Z.~Shi, and Y.~Hu, ``Lightdet: {{A}} lightweight and accurate
  object detection network,'' in \emph{{{ICASSP}} 2020 - 2020 {{IEEE}}
  International Conference on Acoustics, Speech and Signal Processing
  ({{ICASSP}})}, 2020, pp. 2243--2247.

\bibitem{ding2018light}
P.~Ding, Y.~Zhang, W.-J. Deng, P.~Jia, and A.~Kuijper, ``A {{Light}} and
  {{Faster Regional Convolutional Neural Network}} for {{Object Detection}} in
  {{Optical Remote Sensing Images}},'' \emph{ISPRS Journal of Photogrammetry
  and Remote Sensing}, vol. 141, pp. 208--218, 2018.

\bibitem{redmonYouOnlyLook2016a}
J.~Redmon, S.~Divvala, R.~Girshick, and A.~Farhadi, ``You {{Only Look Once}}:
  {{Unified}}, {{Real}}-{{Time Object Detection}},'' in \emph{Proceedings of
  {{IEEE Conference}} on {{Computer Vision}} and {{Pattern Recognition}}}, {Las
  Vegas, NV, USA}, June 2016, pp. 779--788.

\bibitem{wang2020ship}
N.~Wang, B.~Li, X.~Wei, Y.~Wang, and H.~Yan, ``Ship detection in spaceborne
  infrared image {{Based}} on {{Lightweight CNN}} and {{Multisource Feature
  Cascade Decision}},'' \emph{IEEE Transactions on Geoscience and Remote
  Sensing}, pp. 1--16, 2020.

\bibitem{xiaDOTALargeScaleDataset2018}
G.-S. Xia, X.~Bai, J.~Ding, Z.~Zhu, S.~Belongie, J.~Luo, M.~Datcu, M.~Pelillo,
  and L.~Zhang, ``{{DOTA}}: {{A Large}}-{{Scale Dataset}} for {{Object
  Detection}} in {{Aerial Images}},'' in \emph{Proceedings of {{IEEE
  Conference}} on {{Computer Vision}} and {{Pattern Recognition}}}, {Salt Lake
  City, Utah, USA}, June 2018, pp. 3974--3983.

\bibitem{jiang2017r2cnn}
Y.~Jiang, X.~Zhu, X.~Wang, S.~Yang, W.~Li, H.~Wang, P.~Fu, and Z.~Luo,
  ``{{R2CNN}}: {{Rotational Region CNN}} for {{Orientation Robust Dcene Text
  Detection}},'' \emph{arXiv preprint arXiv:1706.09579}, 2017.

\bibitem{nabati2019rrpn}
R.~Nabati and H.~Qi, ``{{RRPN}}: {{Radar Region Proposal Network}} for {{Object
  Detection}} in {{Autonomous Vehicles}},'' in \emph{Proceedings of {{IEEE}}
  International Conference on Image Processing}.\hskip 1em plus 0.5em minus
  0.4em\relax {Taipei, China}: {IEEE}, Sept. 2019, pp. 3093--3097.

\bibitem{xu2020gliding}
Y.~Xu, M.~Fu, Q.~Wang, Y.~Wang, K.~Chen, G.-S. Xia, and X.~Bai, ``Gliding
  {{Vertex}} on {{The Horizontal Bounding Box}} for {{Multi}}-{{Oriented Object
  Detection}},'' \emph{IEEE Transactions on Pattern Analysis and Machine
  Intelligence}, pp. 1--1, Feb. 2020.

\bibitem{qian2019learning}
W.~Qian, X.~Yang, S.~Peng, Y.~Guo, and C.~Yan, ``Learning {{Modulated Loss}}
  for {{Rotated Object Detection}},'' \emph{arXiv preprint arXiv:1911.08299},
  2019.

\bibitem{yang2019condconv}
B.~Yang, G.~Bender, Q.~V. Le, and J.~Ngiam, ``{{CondConv}}: {{Conditionally
  Parameterized Convolutions}} for {{Efficient Inference}},'' in \emph{Advances
  in Neural Information Processing Systems}, vol.~32, 2019, pp. 1307--1318.

\bibitem{tanEfficientDetScalableEfficient2020}
M.~Tan, R.~Pang, and Q.~V. Le, ``{{EfficientDet}}: {{Scalable}} and {{Efficient
  Object Detection}},'' in \emph{Proceedings of {{IEEE Conference}} on
  {{Computer Vision}} and {{Pattern Recognition}}}, {Online}, June 2020, pp.
  10\,781--10\,790.

\bibitem{liObjectDetectionOptical2020}
K.~Li, G.~Wan, G.~Cheng, L.~Meng, and J.~Han, ``Object {{Detection}} in
  {{Optical Remote Sensing Images}}: {{A Survey}} and {{A New Benchmark}},''
  \emph{ISPRS Journal of Photogrammetry and Remote Sensing}, vol. 159, pp.
  296--307, 2020.

\bibitem{linFocalLossDense2017}
T.-Y. Lin, P.~Goyal, R.~Girshick, K.~He, and P.~Dollar, ``Focal {{Loss}} for
  {{Dense Object Detection}},'' in \emph{Proceedings of {{IEEE International
  Conference}} on {{Computer Vision}}}, {Venice, Italy}, Oct. 2017, pp.
  2999--3007.

\bibitem{lawCornerNetDetectingObjects2018}
H.~Law and J.~Deng, ``{{CornerNet}}: {{Detecting Objects}} as {{Paired
  Keypoints}},'' in \emph{Proceedings of the {{European Conference}} on
  {{Computer Vision}}}, {Munich, Germany}, Sept. 2018, pp. 734--750.

\bibitem{chengCrossScaleFeatureFusion2020c}
G.~Cheng, Y.~Si, H.~Hong, X.~Yao, and L.~Guo, ``Cross-{{Scale Feature Fusion}}
  for {{Object Detection}} in {{Optical Remote Sensing Images}},'' \emph{IEEE
  Geoscience and Remote Sensing Letters}, pp. 1--5, 2020.

\bibitem{heMaskRCNN2017b}
K.~He, G.~Gkioxari, P.~Dollar, and R.~Girshick, ``Mask {{R}}-{{CNN}},'' in
  \emph{Proceedings of {{IEEE International Conference}} on {{Computer
  Vision}}}, {Venice, Italy}, Oct. 2017, pp. 2980--2988.

\bibitem{wangPANetFewShotImage2019}
K.~Wang, J.~H. Liew, Y.~Zou, D.~Zhou, and J.~Feng, ``{{PANet}}: {{Few}}-{{Shot
  Image Semantic Segmentation}} with {{Prototype Alignment}},'' in
  \emph{Proceedings of {{IEEE International Conference}} on {{Computer
  Vision}}}, {Seoul, South Korea}, Oct. 2019, pp. 9197--9206.

\bibitem{redmon2017yolo9000}
J.~Redmon and A.~Farhadi, ``{{YOLO9000}}: {{Better}}, faster, stronger,'' in
  \emph{Proceedings of {{IEEE Computer Vision}} and {{Pattern Recognition}}},
  {Honolulu, Hawaii, USA}, June 2017, pp. 7263--7271.

\end{thebibliography}

\end{document}